\pdfoutput=1

\documentclass[11pt]{article}

\usepackage[final]{acl}

\usepackage{times}
\usepackage{latexsym}
\usepackage{kotex}
\usepackage{amsmath}
\usepackage{algorithm}
\usepackage{algpseudocode}
\usepackage{graphicx}
\usepackage{booktabs}
\usepackage{multirow}
\usepackage{array} 
\usepackage{pbox} 
\usepackage{enumitem}
\usepackage{rotating} 

\usepackage[T1]{fontenc}

\usepackage[utf8]{inputenc}

\usepackage{microtype}

\usepackage{inconsolata}

\usepackage{graphicx}

\usepackage{xcolor}

\title{Improving Dialogue State Tracking through\\Combinatorial Search for In-Context Examples}

\author{
Haesung Pyun\textsuperscript{1}\quad
Yoonah Park\textsuperscript{2}\quad
Yohan Jo\textsuperscript{1}\thanks{Corresponding author}\
\\  
\textsuperscript{1}Graduate School of Data Science, Seoul National University \\
\textsuperscript{2}Department of Computer Science and Engineering, Seoul National University
\\
\texttt{\{haesung.pyun, wisdomsword21, yohan.jo\}@snu.ac.kr}
}

\begin{document}
\maketitle
\begin{abstract}
{In dialogue state tracking (DST), in-context learning comprises a retriever that selects labeled dialogues as in-context examples and a DST model that uses these examples to infer the dialogue state of the query dialogue. Existing methods for constructing training data for retrievers suffer from three key limitations}: (1) the synergistic effect of examples is not considered, (2) the linguistic characteristics of the query are not sufficiently factored in, and (3) scoring is not directly optimized for DST performance. {Consequently, the retriever can fail to retrieve examples that would substantially improve DST performance.} To address these issues, we present CombiSearch—a method that scores effective in-context examples based on their combinatorial impact on DST performance. 
{Our evaluation on MultiWOZ shows that retrievers trained with CombiSearch surpass state-of-the-art models, achieving a 20× gain in data efficiency} and generalizing well to the SGD dataset. 
Moreover, CombiSearch attains a 12\% absolute improvement in the upper bound DST performance over traditional approaches when no retrieval errors are assumed. 
This significantly increases the headroom for practical DST performance while demonstrating that existing methods rely on suboptimal data for retriever training.\footnote{Our code: \href{https://github.com/holi-lab/combisearch}{https://github.com/holi-lab/combisearch}}
\end{abstract}

\section{Introduction}
\begin{figure}[t]
  \includegraphics[width=\columnwidth]{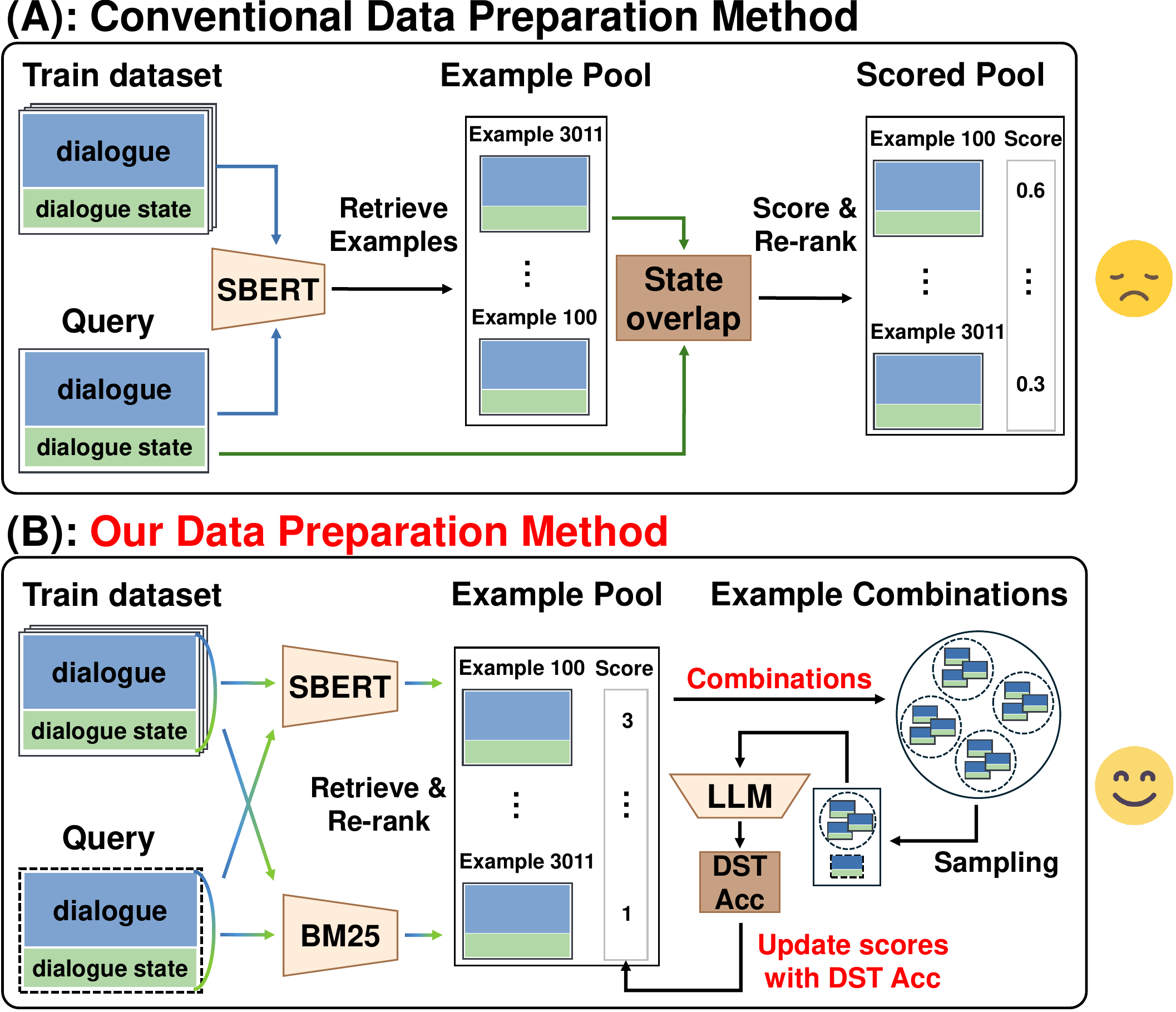}
  \vspace{-0.5em}   \caption{(A) Conventional data preparation method. (B) Our data preparation method that integrates diverse pool construction and combinatorial re-ranking based on DST performance.}
  \label{fig:overview_first_page}
\end{figure}

Dialogue state tracking (DST), a core task for task-oriented dialogue systems and tool agents, tracks user intentions and relevant constraints throughout the dialogue.
For example, if a user says ``I want to book a guest house'' and later adds ``I want to book it for four people'', the dialogue state is updated with the slot-value pairs `[<hotel-type: guest house>, <hotel-book people: 4>]'.

{Recent work shows that large language models can perform DST through in-context learning (ICL) \citep{NEURIPS2020_1457c0d6, zhao2023survey}. By supplying query–response exemplars in the prompt, models learn task-specific patterns from in-context examples, thereby eliminating the need for costly fine-tuning or extensive annotation \citep{NEURIPS2020_1457c0d6, zhang2022opt, JMLR:v24:22-1144, hoffmann2022an, touvron2023llama2openfoundation}. However, since performance depends on the relevance of these exemplars \citep{liu-etal-2022-makes}, recent studies propose retrievers that dynamically select examples tailored to each test query, boosting the accuracy and robustness of ICL performance across tasks \citep{levy-etal-2023-diverse, Lu0CWZRCK23, 10.5555/3618408.3620070}.}

To select effective examples, it is crucial to train an ICL example retriever on high-quality data.
The most widely adopted approach for preparing retriever training data follows two steps \cite{hu-etal-2022-context}: (1) \textbf{pool construction} selects dialogues similar to the target query, and (2) \textbf{re-ranking} prioritizes dialogues whose dialogue states substantially overlap with the query’s dialogue state (Figure~\ref{fig:overview_first_page}-A). However, what constitutes effective in-context examples for DST and how to systematically select them remain understudied and particularly challenging, as DST requires a comprehensive understanding of various conversational aspects \citep{chen-etal-2023-stabilized}.
For instance, coreference resolution is both crucial and error-prone, requiring the system to correctly identify anaphors (e.g., ``it'') and their corresponding antecedents (e.g., ``guest house''), which may span multiple dialogue turns. Our analysis identifies prominent errors and challenges in prior approaches (Appendix~\ref{sec:app_error_analysis}).

We focus on three key limitations in the prior data preparation process and address them in this paper. First, the re-ranking step scores each example individually, which can result in selecting redundant in-context examples that offer little new information. 
Second, relying solely on dialogue state similarity may neglect important contextual information within the dialogue itself and retrieve examples whose conversational characteristics differ significantly from the query dialogue. Such examples are suboptimal for handling complex linguistic phenomena, such as coreference resolution, present in the query dialogue. 
Furthermore, re-ranking examples based on dialogue state similarity does not directly optimize for DST performance, resulting in only ``indirect supervision''.

To address these limitations, we introduce CombiSearch, a new method for preparing training data for in-context example retrievers in DST. By combining diverse pool construction with combinatorial re-ranking (Figure~\ref{fig:overview_first_page}-B), CombiSearch has three key characteristics.
First, the pool construction step collects diverse dialogues that share similar linguistic characteristics with the query---such as conversational flow, style, and structure---beyond just dialogue states. This ensures that more relevant examples are considered during re-ranking, improving the system's ability to handle the linguistic phenomena present in the query.
Second, rather than re-ranking individual examples in isolation, CombiSearch scores examples based on how they function together in tandem. By evaluating their combinatorial impact on DST performance, it accounts for synergistic effects that might be overlooked when scoring examples separately.
Finally, the re-ranking step scores examples by directly optimizing Joint Goal Accuracy (JGA), a standard metric for DST, rather than relying on the indirect signal of dialogue state similarity between each example and the query.

According to our evaluation, CombiSearch outperforms existing state-of-the-art methods on both the MultiWOZ 2.4 and Schema-Guided Dialogue (SGD) datasets. Notably, CombiSearch is highly data-efficient, enabling a retriever trained on CombiSearch-scored data to surpass all baseline models while using only 5\% of the training data required by these models. 
Understanding the upper bound performance achievable with retriever training data \textit{without retriever errors} is crucial, as this sets a hard limit on what practical retrievers can achieve. 
Under this ideal assumption of no retrieval errors, using examples scored by CombiSearch outperformed earlier approaches by 12\% JGA, highlighting that prior methods rely on suboptimal data for training retrievers and that a more effective data preparation approach like CombiSearch is imperative for breakthroughs in DST.
Notably, all these advances from CombiSearch's combinatorial nature come with a running time that is linear in the number of examples to score.

Our key contributions are as follows:
\begin{enumerate}[topsep=2pt, partopsep=0pt, itemsep=1pt, parsep=0pt, leftmargin=15pt]
    \item We present CombiSearch, an effective combinatorial method for constructing data to train retrievers of in-context examples in DST.
    \item We demonstrate that CombiSearch outperforms baselines in DST, even with 20x data efficiency.
    \item We reveal the significantly improved upper bound performance of CombiSearch when no retrieval errors are assumed, highlighting the importance of optimizing retriever training data.

\end{enumerate}

\section{Problem Setting}
\paragraph{Dialogue State Tracking}
DST focuses on predicting the dialogue state that captures the user’s intentions and goals, based on both the current user-agent interaction and the conversation history. 
Let $A^t$ and $U^t$ denote the agent’s and user’s utterances at turn $t$, respectively. 
The dialogue history up to turn $t$ is $C^t = [(A^1, U^1), (A^2, U^2), \dots, (A^t, U^t)]$. 
The objective is to infer a dialogue state $y^t$ from $C^t$, where $y^t$ is a set of slot-value pairs: $y^t = \{(s^1, v^1), (s^2, v^2), \dots, (s^n, v^n)\}$.

Following \citet{hu-etal-2022-context}, we adopt a state-change formulation. 
Specifically, the model takes the previous dialogue state $y^{t-1}$ and current utterances $(A^t, U^t)$ as input, denoted as $x^t = [y^{t-1}, (A^t, U^t)]$, where $y^{t-1}$ summarizes $C^{t-1}$. It predicts the dialogue state change $\Delta y^t = \{+ (s^i, v^i), \dots, - (s^j, v^j)\}$, with ``+'' indicating addition and ``–'' indicating removal of slot-value pairs. The $t$-th dialogue state is $y^t = y^{t-1} \cup \Delta y^t$.

\paragraph{In-Context Learning for DST}
Given a test query $x_{\text{test}}$, the ICL approach retrieves $k$ labeled examples $\{(x_i, \Delta y_i)\}_{i=1}^{k}$ from the training set $D_{\text{train}}= \{(x_i, \Delta y_i)\}_{i=1}^{N}$.
The goal is to predict the dialogue state change of the test query $x_{test}$:
\begin{align}
    \Delta \hat{y}_{\text{test}} \sim P_{\text{LLM}}\bigl(y_\text{test}|\{(x_i, \Delta y_i)\}_{i=1}^{k},\, x_{\text{test}}\bigr)
\end{align}
where $\sim$ refers to a decoding method (e.g., greedy decoding, beam search, etc.).

{In conventional retriever-based ICL for DST, two stages are involved. 
\emph{Example-Retrieval Training} trains the retriever on query-example pairs: each query is paired with candidates ranked by dialogue state similarity, and a contrastive loss draws similar embeddings together while repelling dissimilar ones. 
\emph{Inference Phase} uses the trained retriever to fetch top-ranked examples based on the test query; these examples are prepended to the prompt so the language model can exploit them to generate the query’s dialogue state.}

\section{Method}
\begin{figure*}[t]
    \includegraphics[width=\textwidth]{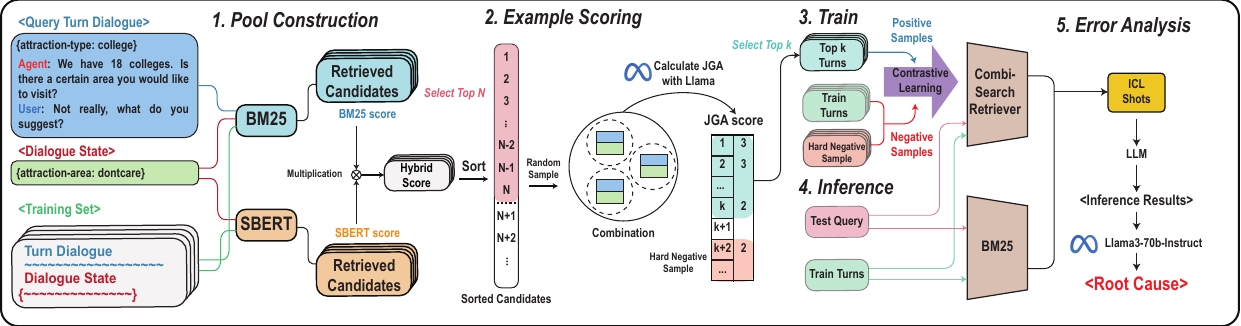}
    \vspace{-1em}   \caption{Overall structure of CombiSearch.}
    \label{fig:combisearch}
\end{figure*}

We present \textbf{CombiSearch}, an algorithm that aims to identify an optimal combination of in-context examples to improve overall DST performance. We first detail each step of CombiSearch (\S{\ref{sec:combisearch}}) and introduce a retriever training method based on CombiSearch-scored data (\S{\ref{sec:combisearch_retriever_training}}). 
Lastly, we describe our text-to-JSON prompt format (\S{\ref{sec:prompting}}).
The overall procedure is illustrated in Figure~\ref{fig:combisearch}.

\subsection{CombiSearch} \label{sec:combisearch}
CombiSearch is designed to generate training data for a retriever of in-context examples for DST. 
The resulting data consists of <query, scored examples> pairs which can be used to train a retriever to select high-scoring examples for a given query. 
CombiSearch unfolds in two complementary steps. First, for each query, it constructs a diverse pool of examples $E$ by leveraging both semantic and lexical similarity to the query. 
Second, it employs a combinatorial mechanism to score the examples based on their collective contribution to successfully predicting the query's dialogue state.
The algorithm is presented in Algorithm~\ref{alg:combisearch}.
Please note that during data preparation and retriever training, each query is drawn from the training set. 

\paragraph{Pool Construction (Figure~\ref{fig:combisearch}-1)} 
Our motivating observation is that DST errors often arise when essential linguistic phenomena required for accurately predicting the query's dialogue state (e.g., coreference resolution) are not adequately represented in the in-context examples. 
Hence, it is important to include diverse examples covering a broader range of linguistic phenomena relevant to the query. 
To that end, we adopt a hybrid search that integrates both semantic and lexical aspects of similarity using two retrievers: SBERT~\citep{reimers-gurevych-2019-sentence} and BM25~\cite{10.1561/1500000019}.
SBERT captures the semantic meaning of dialogue states while BM25 effectively captures linguistic properties---such as word choices, the use of coreferences, named entities, domains, and slots (Appendix~\ref{subsec:app_ret_details_ret_input_comparison}).
We first retrieve the top-$N$ examples from each retriever and merge them into a single list. We then re-rank them based on a hybrid score defined as the product of BM25 TF-IDF score and SBERT cosine similarity score (Appendix~\ref{subsec:app_analysis_hybrid_score}). Finally, we retain the top-$N$ examples to form the final pool $E$.

\begin{algorithm}[t]
\small
\vspace{-0.2em}   \caption{Pseudo-code for \textbf{CombiSearch}}
\label{alg:combisearch}
\begin{algorithmic}[1] 
\State\textbf{Input:} 
\hspace{1mm} $D_{\text{train}}$: Training Dataset, $D_\text{query}$: Query Dataset
\State \textbf{Output:}
\hspace{1mm} $D_{\text{CombiSearch}} = \{(x_i, y_i, E_i)\}_{i=1}^{|Q|}$
\State \textbf{Parameters:}
\hspace{1mm} $k$: Number of in-context examples, $M$: Number of evaluations per example, $P_i$: Input prompt

\vspace{2mm}
\For{$(x_{\text{test}}, y_{\text{test}}) \in D_\text{query}$} \Comment{Pool Construction}
    \State $D_i \gets \text{top-k}_{\text{BM25}}(D_{\text{train}}, N)$
    \State $D_j \gets \text{top-k}_{\text{SBERT}}(D_{\text{train}}, N)$
    \State $D_{\text{union}} \gets D_i \cup D_j$
    \For {$u \in D_{\text{union}}$}
    \State $u.\text{hybrid\_score} \gets u.\text{TF-IDF} \times u.\text{cos\_sim}$
    \EndFor
    \State $E \gets \text{top-k}_{\text{hybrid\_score}}(D_{\text{union}}, N)$
    \For{$i = 1$ \textbf{to} $M$} \Comment{Example Scoring}
        \State $\mathrm{R_i} \gets \text{select\_random\_subset}(E, k)$
        \State $\text{jga}_i \gets \text{JGA}(\text{LLM}(P_i, x_{\text{test}}), y_{\text{test}})$
        \For{$e \in P_i$}
        \State $e.\text{\textit{CombiScore}} \gets  e.\text{\textit{CombiScore}} + \text{jga}_i$
        \EndFor
    \EndFor
    \State $D_{\text{CombiSearch}} \gets D_{\text{CombiSearch}} \cup \{(x_{\text{test}}, y_{\text{test}}, E)\}$
\EndFor

\State \Return $D_{\text{CombiSearch}}$
\end{algorithmic}
\end{algorithm}
\paragraph{Example Scoring (Figure~\ref{fig:combisearch}-2)}
After constructing the candidate pool $E$ with $N$ examples per query, we score individual examples. Rather than scoring an example solely based on its similarity to the query as in existing work, we directly measure its impact on DST performance when used in ICL in tandem with other examples. This approach provides more direct supervision. Our objective is to identify an optimal set of examples that collectively enhance DST performance.

However, exhaustively evaluating all possible combinations is impractical due to the combinatorial explosion.
Instead, we conceptualize each example as a key ``team player'' if it consistently contributes to the success of the ``teams'' it belongs to (i.e., accurate DST) across different sampled combinations. 
CombiSearch implements this by iteratively sampling $k$ examples and evaluating the DST model's prediction using JGA. A JGA score of 1 increases the score of every example in the sampled combination by 1, while a score of 0 leaves the scores unchanged. Repeating this sampling process $M$ times yields an accumulated score for each example---termed \textbf{CombiScore}---reflecting its contribution to DST. 
Notably, this algorithm allows CombiSearch to run in linear time with respect to the number of examples to score while preserving its combinatorial nature. 
The resulting dataset $D_{\mathrm{CombiSearch}}$ pairs each query with a pool of examples $E$ ranked by their CombiScores.

\subsection{Retriever Training (Figure~\ref{fig:combisearch}-3)}\label{sec:combisearch_retriever_training}

We train a specialized retriever to identify effective in-context examples for each query. 
The retriever takes a dialogue (\(y^{t-1}\), \(A^t\), and \(U^t\)) as input and produces an embedding. The training objective is to maximize the embedding similarity between a query and its high-scoring examples in $D_{\mathrm{CombiSearch}}$.

We construct training instances as follows. For each query \(x_i\), we first construct a set of positive examples consisting of the top \(|P|\) examples ranked by CombiScore from pool \(E\); ties are broken by the degree of dialogue state overlap.
Next, for every pair of \(x_i\) and a positive example \(e_j^+\), a set of negative examples is constructed by selecting the bottom \(B\) examples from \(E\) and additional \(B-1\) examples randomly sampled from outside \(E\). As a result, the training data comprise tuples of the form \(\langle x_i, e_j^+, e_{j,1}^-, \dots, e_{j,2B-1}^- \rangle\) for all \(j = 1, ..., |P|\).

Based on this data, we fine-tune our retriever using the InfoNCE loss~\cite{oord2019representationlearningcontrastivepredictive}:
\begin{equation}
\footnotesize 
L(x, P, N) = \sum_{e^+ \in P} -\log \frac{\exp\bigl(\text{sim}(x, e^+)\bigr)}{\sum\limits_{e' \in N \cup \{e^+\}} \exp\bigl(\text{sim}(x, e')\bigr)},
\label{eq:example}
\end{equation}
where \(x\) is the query, \(P\) is the positive set, and \(N\) is the negative set. Additional hyperparameter details are provided in Appendix~\ref{subsec:app_exp_settings_hyperparam}.

At inference time (Figure~\ref{fig:combisearch}-4), for each query, we first retrieve $k$ candidate examples from both our retriever and BM25 and merge the two lists. We then re-rank the merged candidates with the hybrid score introduced in \S\ref{sec:combisearch}, defined as the product of the two retrievers’ scores, and finally select the top $k$ examples as in-context examples. The number of retrieved examples does not have to be equal to the combination size used in CombiSearch.

\subsection{Prompting with Text-to-JSON} \label{sec:prompting}
This section describes our text-to-JSON prompt for DST, which we developed to avoid formatting errors observed when treating DST as a code generation task inspired by \citet{10.1145/3613905.3650756}. 

The prompt begins with a system message that instructs the model to act as a DST expert \citep{kong-etal-2024-better} and provides an explicit task definition \citep{he2024doespromptformattingimpact}. Next, domain-specific details—such as slots, types, and possible values for categorical slots—are provided. In-context examples follow, presented as tuples \( \bigl([y^{t-1}, A^t, U^t], \Delta y^t \bigr) \) in which both the prior state \(y^{t-1}\) and the state changes \(\Delta y^t\) are represented in JSON format. The prompt ends with clear DST instructions and output format specifications, all formatted in Markdown for clarity (see Appendix~\ref{sec:app_prompt}).

\begin{table*}[hbt!]
  \centering
  \resizebox{\textwidth}{!}{ 
    \tiny
    \begin{tabular}{p{0.05cm} p{3.4cm}|l|ccc|cccc}
      \hline
      & \multirow{2}{*}{Method} & \multirow{2}{*}{}{Language} & \multicolumn{3}{c|}{MultiWOZ 2.1}& \multicolumn{3}{c}{MultiWOZ 2.4}\\
        &                     &{Model} & 1\%        & 5\%         & 100\%       & 1\%         & 5\%         & 100\%          \\
      \hline
      \multirow{4}{*}{\rotatebox[origin=c]{90}{Fine-tuned}} & DiSTRICT\citep{venkateswaran-etal-2023-district} & T5-small       & 13.4       & 41.3        & 56.1        & -           & -           & -            \\
      & SGPDST\citep{lee-etal-2021-dialogue} & T5-base        & 32.1       & 43.1        &\textbf{56.7}& -           & -           & -            \\
      & SM2-11B\citep{chen-etal-2023-stabilized} & T5-XXL      & 38.4	    & 44.6	     & -           & 40.0	      & 51.1	    & -           \\
      & DS2\citep{shin-etal-2022-dialogue}     & T5-large        & 33.8       & 44.2        & 53.3        & 36.8        & 49.9        & 57.9         \\     
      \hline
      \multirow{4}{*}{\rotatebox[origin=c]{90}{ICL}} & IC-DST\citep{hu-etal-2022-context}  &\texttt{code-davinci-002}  & 43.1	    & 47.1	      & 48.7	    & 50.7	      & 48.4	    &55.4         \\
      & SynthDST\citep{kulkarni-etal-2024-synthdst}&\texttt{gpt-3.5-turbo}  &\textbf{45.8}& 44.9        & 46.0        & 51.0        & 50.4        & 55.2        \\
      & RefPyDST*\citep{king-flanigan-2023-diverse}&\texttt{gpt-3.5-turbo}  & 39.7        & 43.7        & 48.9        & 44.9        & 52.3        & 58.0        \\
      & CombiSearch&\texttt{gpt-3.5-turbo}&\textbf{45.8}&\textbf{47.7}& 50.0        &\textbf{56.7}&\textbf{59.8}& \textbf{64.2}\\
      \hline
    \end{tabular}
  }
  \vspace{-0.5em}  
  \caption{
  JGA scores in the closed-source setting, evaluated on the full test sets of MultiWOZ 2.1 and 2.4 using 1\%, 5\%, and 100\% of the training data. Each score represents the mean of three runs. *RefPyDST was reproduced under our own settings because \texttt{code-davinci-002} has been deprecated and the log-probabilities of input prompts are no longer provided by the API. More details are provided in Appendix~\ref{subsec:app_exp_settings_baseline}.}
  \label{tab:few_shot_dst}
\end{table*}

\section{Experiments}
In this section, we conduct ICL experiments for DST in various settings. The practical setting (\S{\ref{sec:exp_practical}}) uses a trained retriever and evaluates DST performance on in-domain data (\S{\ref{sec:exp_seen}}), unseen data (\S{\ref{sec:exp_general}}), and a dataset that involves coreference resolution (\S{\ref{sec:ablation_coref}}). 
On the other hand, the oracle setting (\S{\ref{sec:exp_upperbound}}) examines the upper bound DST performance without retrieval errors. 
In ablation studies (\S{\ref{sec:ablation_pool_combi}}), we assess the contribution of individual components in CombiSearch. Furthermore, we compare the combinatorial scoring method with an individual scoring baseline to measure the effect of modeling interactions among examples.

Common experimental settings are described below, with specific configurations detailed in their respective subsections. Further analysis of the experimental settings can be found in Appendix~\ref{sec:app_analysis}.

\paragraph{Common Experiment Settings}\label{sec:exp_settings}
We evaluated our method on the MultiWOZ and Schema-Guided Dialogue (SGD) datasets. The MultiWOZ dataset comprises roughly 10,000 human-human dialogues across five domains (hotel, attraction, taxi, train, and restaurant) for training, with evaluation conducted on MultiWOZ 2.4. The SGD dataset contains 16,000 machine-machine dialogues across 26 services, with unseen domains in its test set used to assess generalization.
Performance is measured using Joint Goal Accuracy (JGA), which requires exact matches between predicted and ground-truth dialogue state changes at each turn. 

Our CombiSearch configuration uses a candidate pool size of \(N=100\) and \(k=10\) examples in each combination. We set \(M=3\) evaluations per example to balance accuracy with computational efficiency on large-scale datasets (Appendix~\ref{subsec:app_analysis_num_eval}). {During CombiSearch, we score candidate examples with \texttt{Llama-3-8B-Instruct} as the DST model. At inference time, we experiment with \texttt{Llama-3-8B-Instruct} for the DST model. To assess data efficiency, we explore using 1\%, 5\% and 100\% of the training data, for both data construction for retriever training during CombiSearch and data retrieval at inference time. Further implementation details are in Appendix~\ref{sec:app_exp_settings}.}

\subsection{Performance with Trained Retrievers}\label{sec:exp_practical}

We evaluate CombiSearch's DST performance using retrievers fine-tuned on \(D_\text{CombiSearch}\). 

\subsubsection{In-domain Performance}\label{sec:exp_seen}
We cover two scenarios: a \textbf{Closed-source} setting using black-box LLMs and an \textbf{Open-source} setting where only open-source LLMs are available.

\paragraph{Settings}
The \textbf{Closed-source} setting is as follows. When we explore using only 1\% and 5\% of the training data, we run experiments three times with random samples of the training data (following \citet{hu-etal-2022-context}).
Unlike the common practice of training separate retrievers for each split \citep{hu-etal-2022-context, king-flanigan-2023-diverse}, however, we train a single retriever on one split and use it for all three splits, which serve as retrieval corpora. This out-of-distribution (OOD) setup is more challenging than using a retriever trained specifically for each split, as the retriever must select effective examples from unseen data during inference. We use \texttt{gpt-3.5-turbo} as the DST model and report the mean performance across the three runs.

For \textbf{Open-source} experiments, we sample 1\% and 5\% of the training data once, which serve as both the retriever training data and the retrieval corpus. {We experiment with \texttt{Llama-3-8B-Instruct} and a 4-bit GPTQ-quantized \texttt{Llama-3-70B-Instruct} (to accommodate resource constraints) as the DST model. 
Details about the baseline models used in the experiments are provided in Appendix~\ref{subsec:app_exp_settings_baseline}.} 

\paragraph{Results}
{For the \textbf{Closed-source} setting, Table~\ref{tab:few_shot_dst} shows that CombiSearch outperforms all baselines in data-constrained settings (1\% and 5\%) on both MultiWOZ 2.1 and 2.4. Notably, CombiSearch with 5\% of training data even surpasses every baseline trained on the full dataset on MultiWOZ 2.4. Furthermore, with only 1\% of training data, CombiSearch still exceeds SynthDST, which is trained on its full synthetic data. These results highlight the effectiveness and data efficiency of CombiSearch.}

\begin{table}[t]
  \centering
  \resizebox{\columnwidth}{!}{
    \begin{tabular}{l|l|ccc}
    \hline
    \textbf{Retrieval} & \textbf{DST Model} & \textbf{1\%} & \textbf{5\%}   & \textbf{100\%} \\
    \hline
    RefPyDST                & \texttt{Llama-3-8B-Instruct}               & 47.7         & 50.6          & 55.5          \\ 
    CombiSearch             & \texttt{Llama-3-8B-Instruct}               & \textbf{52.1}& \textbf{56.2} & \textbf{61.8} \\\hline
    RefPyDST                & \texttt{Llama-3-70B-Instruct}              & 56.0         & 57.1          & 56.2          \\
    CombiSearch             & \texttt{Llama-3-70B-Instruct}              & \textbf{56.6}& \textbf{59.3} & \textbf{61.5}          \\
    \hline\hline
    RefPyDST                & \texttt{gpt-3.5-turbo}          & 46.2         & 52.7          & 58.0          \\
    CombiSearch             & \texttt{gpt-3.5-turbo}          & 56.6         & 59.3          & 64.2          \\\hline
    \end{tabular}
  }
  \vspace{-0.5em}   
  \caption{{JGA scores in the open-source setting on MultiWOZ 2.4, using 1\%, 5\%, and 100\% of the training data. \texttt{gpt-3.5-turbo} performance on the same training data is also provided.}}
  \label{tab:open_colse_comparison}
\end{table}

{In the \textbf{Open-source} setting (Table \ref{tab:open_colse_comparison}), CombiSearch exceeds RefPyDST by more than 5\% in JGA when \texttt{Llama-3-8B-Instruct} serves as the DST model, regardless of the amount of training data. A consistent---though smaller---average gain about 3\% appears with the 4-bit–quantized \texttt{Llama-3-70B-Instruct}. 
The last two rows of the table show the performance of \texttt{gpt-3.5-turbo} when the same training samples are used for a fair comparison, which exhibit the consistent tendency.}

{Although constructing CombiSearch data for training a retriever takes more wall-clock time than RefPyDST for the same training data (Appendix \ref{sec:app_exp_settings_computing}), CombiSearch achieves equal or higher JGA in less overall time because it requires scoring far fewer training data. With only 5\% of the training data, CombiSearch reaches 59.3\% JGA, already surpassing RefPyDST trained on the full training data (56.2\% JGA). 
Moreover, the running time of CombiSearch increases only linearly with the number of examples to score, and this is only a one-time cost for retriever training that does not affect inference time. 
Therefore, in practical deployment settings where a retriever is updated only occasionally, the strong data efficiency and performance gains from CombiSearch compensate for its data construction cost.}


\subsubsection{Generalization Ability}\label{sec:exp_general}
{To verify that CombiSearch’s advantage generalizes beyond MultiWOZ, we additionally evaluate on SGD dataset. We consider two experimental settings: (1) \emph{in-domain}, where a retriever is trained and tested on SGD, and (2) \emph{cross-dataset}, where a retriever trained on MultiWOZ is applied to SGD without adaptation.}

\paragraph{Settings}  
{We train two retrievers on 5\% of the training data: one on SGD and the other on MultiWOZ 2.1. Both were evaluated on 30\% of the SGD test set. During inference, in-context examples were drawn from the same 5\% subset of SGD, for a fair comparison. DST performance was measured with JGA for both the cumulative state ($y_{\text{test}}$) and the per-turn state change ($\Delta y_{\text{test}}$). The \textbf{Random} baseline selects examples randomly. \texttt{Llama-3-8B-Instruct} serves as the DST model.}

\begin{table}[t]
\centering
\resizebox{\columnwidth}{!}{
\begin{tabular}{l|l|ccc}
\hline
\multirow{2}{*}{\textbf{\shortstack{Training\\Dataset}}} & \multirow{2}{*}{\textbf{Metric}} & \multicolumn{3}{c}{\textbf{Retrieval Method}} \\ 
\cline{3-5}
& & \textbf{Random} & \textbf{RefPyDST} & \textbf{CombiSearch}\\
\hline
\multirow{2}{*}{SGD} 
    & $y_{\text{test}}$         & 4.45 & 15.81 & 16.81 \\
    & $\Delta y_{\text{test}}$  & 37.25 & 53.17 & 54.41 \\
\hline
\multirow{2}{*}{MultiWOZ} 
    & $y_{\text{test}}$         & 4.45  & 12.08 & 12.24 \\
    & $\Delta y_{\text{test}}$  & 37.25 & 49.16 & 50.16 \\
\hline
\end{tabular}}
\vspace{-0.5em}
\caption{{JGA scores on the SGD test set. For “SGD” and “MultiWOZ” rows, retrievers were trained on 5\% of SGD and MultiWOZ 2.1, respectively.}}
\label{tab:sgd_performance}
\end{table}

\paragraph{Results}  
{Table \ref{tab:sgd_performance} shows that both CombiSearch and RefPyDST generalize beyond their training corpus, outperforming the Random baseline by large margins. When the retriever is trained \emph{in-domain (SGD)}, CombiSearch attains 54.41\% JGA on $\Delta y_{\text{test}}$, edging out RefPyDST by roughly one 1\% JGA and achieving a similar improvement on $y_{\text{test}}$ (16.81\%). The same pattern holds under the \emph{cross-dataset} setting: CombiSearch trained on MultiWOZ still surpasses its RefPyDST counterpart on both metrics. Though modest, this consistent edge under both domain-matched and domain-shift conditions underscores that CombiSearch is not merely overfitting to a single corpus. Rather, it preserves a performance lead across different domains.}


\subsubsection{Coreference Resolution}\label{sec:ablation_coref}

We evaluate the effectiveness of CombiSearch in dialogues that require coreference resolution, a critical challenge in multi-turn, cross-domain DST. Because speakers routinely employ ellipsis or pronominalization, a system must resolve such contextual references and extract precise slot values from ongoing dialogues.

\paragraph{Settings}
Experiments are conducted on MultiWOZ 2.3, which provides turn-level coreference annotations \citep{han2021multiwoz23multidomaintaskoriented}. Following \citet{king-flanigan-2023-diverse}, we compute (i) slot-level accuracy on turns that require coreference resolution and (ii) JGA restricted to dialogues that contain at least one such turn. For few-shot evaluation, we use 5\% of the training data both to train the retriever and as the retrieval corpus. Two retrievers---RefPyDST and CombiSearch---are compared. 

In this experiment, our focus is primarily on examining the differences in retrieved examples between the two methods. Therefore, we control for several factors that differ between them by using only embedding-based retrieval (instead of hybrid retrieval) and excluding the diversity-promoting re-ranking employed by RefPyDST.  
Since they use different prompt formats---Text-to-JSON (`JSON') and Text-to-Python (`Python')---we further control for prompt format. 
In the zero-shot condition (0\%), we follow \citet{hu-etal-2022-context} and provide a single example to inform the output format. \texttt{Llama-3-8B-Instruct} serves as the DST model.

\paragraph{Results}
{Table \ref{tab:coreference_performance} shows that, in the 5\% setting, CombiSearch surpasses RefPyDST on accuracy and JGA by up to 2.7\% (JSON 83.6\% vs 81.6\%; Python 23.6\% vs 21.3\%).} This suggests that CombiSearch retrieves examples that are more helpful for coreference resolution than RefPyDST. In the zero-shot setting, the `JSON` format yields better performance than the `Python` format. This improvement may be due to frequent formatting errors in the `Python` prompt, as well as the superiority of our formatting instructions.

When a query requires coreference resolution, CombiSearch retrieved such examples slightly more often than RefPyDST (3.7 vs. 3.6 per query) and achieved about 2\% higher JGA than RefPyDST (72.4\% vs. 70.7\%). These findings further highlight the strength of CombiSearch in retrieving effective examples for handling linguistic phenomena such as coreference resolution. Representative cases are provided in Appendix~\ref{sec:app_coref-ex}. 

\subsection{Upper Bound without Retrieval Errors}\label{sec:exp_upperbound}
In this section, we run experiments to establish the \textit{oracle} upper bound performance of CombiSearch by removing retrieval errors. Specifically, rather than using retrievers at inference time, we directly select the highest-scoring examples according to CombiScore with access to the query's ground-truth dialogue state. Although this setting is impractical in real-world applications, where the ground-truth dialogue states of queries are unavailable, it serves two key purposes: (i) estimating the maximal DST performance achievable from the retriever training data and (ii) assessing how effectively existing DST methods leverage available data.

Additionally, we conduct an error analysis using DST-EA, a tool we developed to categorize errors into seven types and twenty subclasses. This enables a deeper understanding of DST model behavior and contributes to further advancements in DST research. A detailed description of DST-EA and the error types is provided in Appendix~\ref{sec:app_error_analysis}.

\paragraph{Settings}  
Due to computational constraints, we use three separate 10\% subsets of the test set and report the mean performance across these subsets to ensure statistical reliability. For each test query, we build \(D_{\text{CombiSearch}}\) from the training set, exploring the use of either 1\% or 5\% of the data to examine the impact of example diversity on DST performance. 
We compare our method against two baselines: RefPyDST and a “Hybrid” approach that employs the pool construction process in CombiSearch and ranks examples based on the hybrid score (from BM25 and SBERT) without applying combinatorial scoring. For fair comparison, RefPyDST and Hybrid also skip retrieval errors and are given access to the ground-truth dialogue states of test queries. We employ both \texttt{Llama-3-8B-Instruct} and the 4-bit–quantized \texttt{Llama-3-70B-Instruct} as the DST model. 

\begin{table}
  \small
  \centering
  \resizebox{\columnwidth}{!}{ 
  \begin{tabular}{c|c|c|c|c|c}
    \hline
    \multirow{2}{*}{\textbf{Prompt}} & \multirow{2}{*}{\textbf{Retriever}} & \multicolumn{2}{c|}{\textbf{5\%}} & \multicolumn{2}{c}{\textbf{0\%}} \\
    && \textbf{Acc}   & \textbf{JGA}    & \textbf{Acc}   & \textbf{JGA} \\
    \hline
        \multirow{2}{*}{JSON}        & RefPyDST      & 81.6              & 39.9              & \multirow{2}{*}{48.7}     & \multirow{2}{*}{12.2}\\
                & CombiSearch   & \textbf{83.6}     & \textbf{42.6}     &     & \\
        \hline
        \multirow{2}{*}{Python}      & RefPyDST      & 21.3              & 36.0              & \multirow{2}{*}{15.2}     & \multirow{2}{*}{19.5}\\
              & CombiSearch   & \textbf{23.6}     & \textbf{39.4}     &     & \\
    \hline
  \end{tabular}
  }
  \vspace{-0.5em}   
  \caption{Performance on coreference resolution data. \textbf{Acc}: the accuracy of slots for which coreference resolution is required. \textbf{JGA}: JGA for dialogue turns that require coreference resolution.}
  \label{tab:coreference_performance}
\end{table}

\paragraph{Results}
As shown in Table~\ref{tab:oracle_performance}, CombiSearch achieves 82.7\% JGA with \texttt{Llama-3-8B-Instruct} (row 3) and 79.6\% with the 4-bit–quantized \texttt{Llama-3-70B-Instruct} (row 6). These scores represent an improvement of roughly 13\% points over RefPyDST for each model (rows 1 and 4). The higher upper bound signifies greater headroom for the practical setting of using trained retrievers. It also indicates that the existing methods relied on suboptimal data for retriever training. 
Furthermore, CombiSearch outperforms the Hybrid baseline by 8\%--12\% (rows 2 and 5), highlighting the importance of aligning the scores of examples with DST performance through a combinatorial manner.

\paragraph{Error Analysis}
CombiSearch markedly reduces the most common errors made by BM25- and SBERT-based systems---chiefly, missing explicit user confirmations and over-interpreting ambiguous user utterances---by modeling context more comprehensively. Notable improvements were observed in recognizing confirmation utterances across multiple turns and achieving more accurate utterance interpretation.

\begin{table}[t]
  \small
  \centering
  \begin{tabular}{ll|cccc}
    \hline
      \textbf{Scoring Method}  & \textbf{Model Size}   & \textbf{1\%}  & \textbf{5\%}      & \textbf{100\%}  \\
      \hline
      RefPyDST              & 8B             & 58.0        & 62.7        & 69.7         \\
      Hybrid                & 8B             & 60.4        & 68.0        & 75.9         \\
      \textbf{CombiSearch}  & 8B             &\textbf{ 68.4}        &\textbf{ 75.1}        & \textbf{82.7}         \\
      \hline
      RefPyDST              & 70B            & 63.8        & 66.4        & 67.9         \\
      Hybrid                & 70B            & 60.2        & 62.3        & 67.6         \\
      \textbf{CombiSearch}  & 70B            & \textbf{71.8}        & \textbf{72.7}        & \textbf{79.6}        \\
      \hline
  \end{tabular}
  \vspace{-0.5em}   \caption{Oracle JGA scores across different scoring methods and models.
}
  \label{tab:oracle_performance}
\end{table}

\subsection{Ablation Study}\label{sec:ablation_pool_combi}
We evaluated the contributions of pool construction and combinatorial search through component ablation. In this analysis, we not only assess the significance of each component but explore the best way to integrate these two steps to maximize DST performance (see Appendix~\ref{sec:app_analysis} for further analysis). 

\paragraph{Settings} 
All the experiments use the oracle setting as in \S{\ref{sec:exp_upperbound}}, accessing test queries’ ground-truth dialogue states and directly selecting the highest-scoring examples. 
We use 5\% of the training data and \texttt{Llama-3-8B-Instruct} as the DST model.

\paragraph{Pool Construction}
The original pool construction step of CombiSearch employs both BM25 and SBERT to retrieve example candidates. To analyze the contribution of each of these retrievers, we explore several configurations and measure DST performance.

As shown in Table~\ref{tab:pool_combinatorial_performance}, using both BM25 and SBERT (row 3) yielded a 14\% higher JGA score than composing the pool with random examples (row 1). Additionally, using SBERT alone (row 2) led to a performance drop of 3\%-4\% compared to the combined approach. This finding underscores the importance of example diversity within the pool. Employing both retrievers results in examples that capture a wider range of dialogue characteristics and enables the combinatorial scoring step to effectively consider and recognize useful examples.

\paragraph{Combinatorial Search}

We compare CombiSearch with the another combinatorial search method $\text{Se}^2$ \cite{liu-etal-2024-se2}, which greedily selects the example that maximally enhances ICL performance when combined with previously selected examples.
For a fair comparison, we use the same three-shot setting as $\text{Se}^2$ ($k = 3$). 
While $\text{Se}^2$ employs a pool size of 50 examples, we use a smaller pool size of 30 ($N=30$) for CombiSearch, which serves as a handicap.
See Appendix~\ref{subsec:app_exp_settings_baseline} for details about ${Se}^2$.

As shown in Table~\ref{tab:pool_combinatorial_performance}, CombiSearch outperformed $\text{Se}^2$ by 12\% (rows 3 and 4). At the same time, CombiSearch is also more cost-efficient, requiring only 30 evaluations to construct $D_{\text{CombiSearch}}$ for a given query (i.e., 30 examples divided into 10 groups and evaluated 3 times), compared to the 150 evaluations needed by $\text{Se}^2$ to select three examples.
The results demonstrate that while each component provides distinct benefits, their optimal integration substantially enhances the overall effectiveness of CombiSearch.

\paragraph{Error Analysis}
We observed a notable behavior for the DST model that was not present in the 10-shot ICL setting. The model initially predicted all domain-related slots as \textit{dontcare} in the first turn before updating them in later turns. This suggests that, with fewer examples, the model overinterprets unspecified slot values as a lack of user preference.

\begin{table}[t]
  \centering
  \small{ 
  \begin{tabular}{>{\centering\arraybackslash}p{2.1cm} >{\centering\arraybackslash}p{2.1cm} >{\centering\arraybackslash}p{1cm} >{\centering\arraybackslash}p{0.8cm}}
    \hline
    \textbf{Pool Construction} & \textbf{Combinatorial Search} & \textbf{\# of Inference} &   \textbf{JGA (\%)} \\
    \hline
        Random          & CombiSearch       & 31      & 56.9   \\
        SBERT           & CombiSearch	    & 31      & 67.7   \\
        BM25\&SBERT     & CombiSearch       & 31      & \textbf{70.7}   \\\hline
        Random	        & N/A               & 1       & 19.2    \\        
        SBERT	        & N/A               & 1       & 60.8    \\
        BM25\&SBERT	    & N/A               & 1       & 58.4   \\\hline
        -    	        & $\text{Se}^2$     & 151     & 58.8    \\
    \hline
  \end{tabular}
  }
  \vspace{-0.5em}   \caption{{Oracle JGA scores across different pool-construction and combinatorial search methods.}}
  \label{tab:pool_combinatorial_performance}
\end{table}

\subsection{Individual vs.\ Combinatorial Scoring}\label{subsec:exp_analysis_combi_vs_indiv}

{We test whether scoring examples in groups (CombiSearch) yields higher accuracy than scoring each example independently ($k=1$ in CombiSearch).}

\paragraph{Setting}
{%
In the oracle setting, for every query, we draw 100 candidate dialogues from the full MultiWOZ training set (using ground-truth dialogue states) and score them with (i) \textit{Individual} (100 calls/query), (ii) \textit{CombiSearch} ($M{=}3$, 30 calls), and (iii) \textit{CombiSearch} ($M{=}9$, 90 calls). In the practical setting, we use 5\% of the training data under the open-source regime described in \S{\ref{sec:exp_practical}}. Example scoring employs \texttt{Llama-3-8B-Instruct} as the DST model, and both \texttt{Llama-3-8B-Instruct} and 4-bit–quantized \texttt{Llama-3-70B-Instruct} are tested during inference.%
}

\paragraph{Results}
{%
In the oracle setting (Table~\ref{tab:oracle_combo_vs_indiv}) CombiSearch with $M{=}3$ improves JGA by 2.8\% over Individual while cutting compute to one-third. With $M{=}9$, the gain widens to 5.4\% at a comparable budget. 
The practical setting mirrors this pattern (Table~\ref{tab:retriever_combo_vs_indiv}): retrievers trained on CombiSearch data surpass their counterparts by roughly 3\%, irrespective of the DST model's size. 
Hence, exploiting interactions among examples not only reduces computational cost but also results in higher-quality training data for retrievers.}
\begin{table}[t]
\centering
  \resizebox{\columnwidth}{!}{ 
    \begin{tabular}{l c c c}
    \hline
    \textbf{Scoring Method} & \textbf{Model} & \textbf{JGA (\%)} & \textbf{Calls/query}\\ \hline
    Individual (100 cand.) & 8 B & 79.9 & 100 \\
    CombiSearch ($M{=}3$)  & 8 B & 82.7 &  30 \\
    CombiSearch ($M{=}9$)  & 8 B & 85.3 &  90 \\ \hline
    \end{tabular}
}
\vspace{-0.7em}
\caption {Oracle JGA scores on the MultiWOZ test set across varying scoring methods.}
\label{tab:oracle_combo_vs_indiv}
\end{table}

\section{Related Work}\label{sec:related_work}

\paragraph{Dialogue State Tracking}
To ensure efficient domain transfer in DST, zero-shot methods have been explored \citep{wu-etal-2019-transferable, 10.5555/3495724.3497418, gupta-etal-2022-show}. But since they struggle when domains differ significantly, few-shot DST has been studied, where models are often fine-tuned to follow in-context demonstrations \citep{shin-etal-2022-dialogue, venkateswaran-etal-2023-district}, with additional approaches exploring instruction tuning \citep{feng-etal-2023-towards} and meta-training \citep{chen-etal-2023-stabilized}. However, these approaches require extensive training and annotation as the DST schema changes.

Another approach is to train retrievers to select in-context examples per query. \citet{hu-etal-2022-context} labeled positive and negative examples and fine-tuned a retriever with the data. They proposed representing a dialogue history as a summary state and tracking state changes at each turn. Expanding on this, \citet{king-flanigan-2023-diverse} applied maximum marginal relevance to diversify retrieved examples, while \citet{kulkarni-etal-2024-synthdst} proposed synthetic data generation for fine-tuning retrievers.

\paragraph{In-Context Example Selection}
Traditional methods often examples based on their impact on task performance \citep{nguyen2023incontextexampleselectioninfluences,chang-jia-2023-data}. 
However, many studies use a fixed set of examples across different queries, limiting query relevance. 
To address this, other works train retrievers to select examples per query \citep{hu-etal-2022-context, king-flanigan-2023-diverse, rubin-etal-2022-learning}. 
Our work presents a data construction method for training retrievers in this setting. 

For data construction, studies have explored scoring functions to distinguish positive and negative examples. Common methods include measuring semantic similarity to the query, applicable to DST via dialogue state similarity \citep{liu-etal-2022-makes, hu-etal-2022-context, king-flanigan-2023-diverse}. Other methods leverage the model's generation probabilities \citep{rubin-etal-2022-learning,iter-etal-2023-context, peng-etal-2024-revisiting}. These generation probability-based methods suffer from surface-form competition \citep{holtzman-etal-2021-surface}. To mitigate this, while accounting for the model's impact, we optimize for end-task performance instead. 
Furthermore, some studies employ combinatorial search as in our work \citep{10.5555/3666122.3669422, gupta-etal-2023-coverage}. For instance, \citet{liu-etal-2024-se2} suggest iteratively selecting the best example at a time in a greedy manner. We demonstrated that our method achieves a lower computational cost and higher effectiveness.

\begin{table}[t]
\centering
\tiny
  \resizebox{\columnwidth}{!}{
    \begin{tabular}{l c c }
    \hline
    \textbf{Scoring Method} & \textbf{Model Size} & \textbf{JGA (\%)} \\
    \hline
        Individual scoring    & 8 B  & 54.3 \\
        CombiSearch ($M{=}3$) & 8 B  & 56.2 \\\hline
        Individual scoring    & 70 B & 56.9 \\ 
        CombiSearch ($M{=}3$) & 70 B & 59.3 \\\hline
    \end{tabular}
    }
    \vspace{-0.7em}
\caption{{Oracle JGA scores by individual and combinatorial scoring (5\%  training data).}}
\label{tab:retriever_combo_vs_indiv}
\end{table}

\section{Conclusion}
We presented CombiSearch, a combinatorial method for scoring in-context examples in DST. We explored the oracle upper bound of performance and demonstrated that CombiSearch-scored examples achieve a substantially higher JGA score than the SOTA model. In practical settings, the retriever trained on CombiSearch-scored examples also outperformed the SOTA model. Our error analysis tool sheds light on common error types made by LLMs. Our work highlights the importance of utilizing high-potential data for retriever training and identifies several areas for improvement in LLMs to reduce common errors.

\section{Limitations}
Our method facilitates effective example selection and training data preparation for DST retrievers. It also demonstrates high ideal performance, defining the upper bound for retriever training. This highlights its potential and suggests that existing methods rely on suboptimal data preparation in DST.

Applying in-context example selection to DST inevitably introduces computational overhead—a well-known challenge in such methods. Since existing approaches require evaluating thousands of candidates against a small example set, computational cost remains a major concern. To address this, we optimized efficiency while ensuring selection aligned with task objectives. As a result, we achieved strong discriminative power with only 30 evaluations on a large example set. Despite these optimizations, high computational overhead remains a key challenge for the field.

Additionally, practical retriever training did not fully reach the estimated upper bound. Our research primarily focuses on example selection and data preparation rather than optimizing retriever training itself. While designing advanced training methodologies is beyond our scope, recent studies have explored reinforcement learning-based iterative retrieval and contrastive learning for ranking. Future work could integrate such approaches to improve retriever performance. Advancements in this direction would not only enhance DST but also refine industrial applications.

\section{Ethical and Social Impacts}

\paragraph{Social Implications} 
Our approach enhances dialogue state tracking efficiency and accuracy by optimizing the retrieval process via CombiSearch. This reduction in computational cost renders high-performing DST systems more accessible, particularly in resource-constrained settings, and promises to improve user experiences in domains such as customer service, healthcare, and education.

\paragraph{Potential Risks and Challenges} 
Despite these benefits, risks remain. If the retrieval process selects irrelevant or low-quality examples, the DST system may yield misleading outputs, potentially propagating misinformation. Moreover, while our method is more efficient than traditional combinatorial approaches, the need for multiple evaluations still poses scalability challenges—issues common to systems leveraging LLMs and example selection methods.

\paragraph{Mitigation Strategies} 
To address these challenges, future work should focus on enhancing retriever robustness to consistently select high-quality, diverse examples, thereby reducing misinformation and improving domain generalization. Additionally, developing techniques to adapt to unseen or out-of-domain dialogue scenarios—such as augmenting training data or designing retrievers to prioritize variability—will be essential. Investigating adaptive evaluation strategies that minimize computational overhead without compromising performance is also a key direction.

By systematically addressing these concerns, our method aims to maximize positive impact while mitigating inherent risks, thus advancing the field of DST.

\section*{Acknowledgments}

This work was supported by the National Research Foundation of Korea (NRF) grant (RS-2024-00414981), by the Institute of Information \& Communications Technology Planning \& Evaluation (IITP) under the Leading Generative AI Human Resources Development grant (IITP-2025-RS-2024-00397085), and by the IITP grant (RS-2025-02215122, Development and Demonstration of Lightweight AI Model for Smart Homes), all of which were funded by the Korean government (MSIT). This research was also supported by the Korea Institute of Science and Technology Information (KISTI) in 2025 (No.(KISTI) K25L1M1C1), aimed at developing KONI (KISTI Open Neural Intelligence), a large language model specialized in science and technology.

\bibliography{custom}

\appendix

\section{Error Analysis Framework}\label{ea tool}

We introduce a comprehensive error analysis tool that categorizes DST prediction errors into seven types, allowing us to better understand the underlying causes of mistakes made by LLMs and inform directions for improving the methods.

\begin{figure*}[t]
  \includegraphics[width=\linewidth]{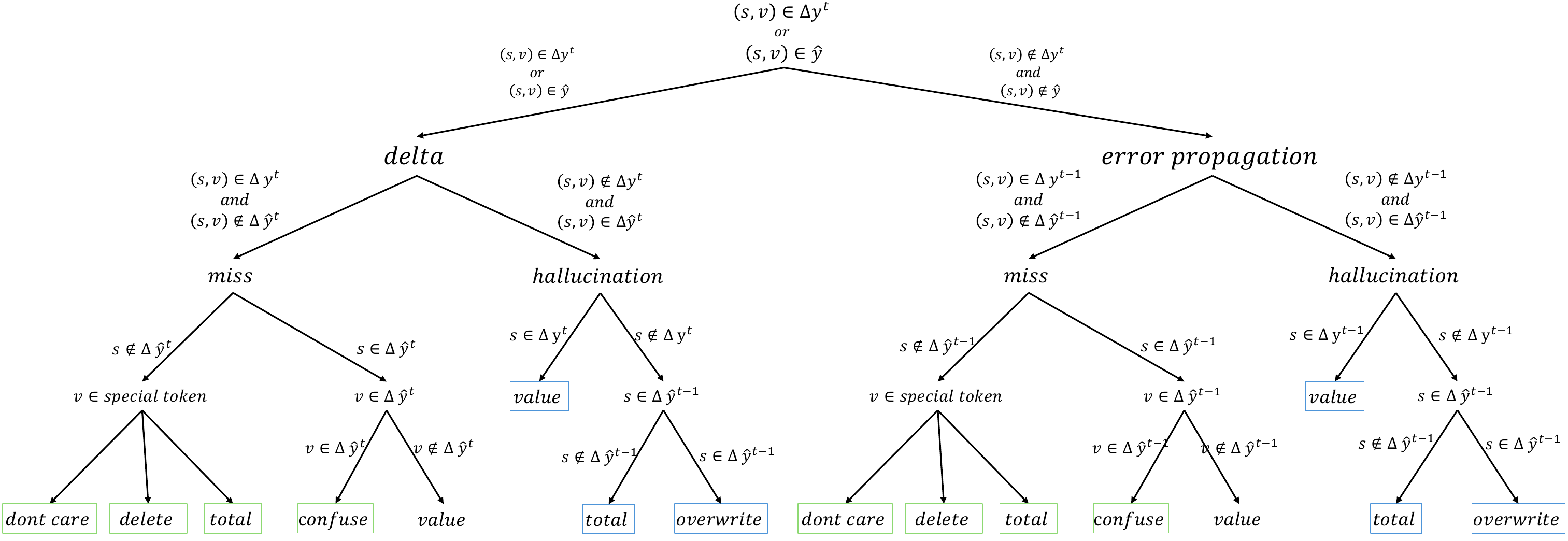}
  \vspace{-0.5em}   \caption{Error tree to divide error type by slot and value}
  \label{fig:errortree}
\end{figure*}

Figure~\ref{fig:errortree} shows our classification of error types (error types are marked in boxes). Recall that the predicted dialogue state in the current turn $t$ is $y^{t} = \Delta y^t \cup y^{t-1}$. Hence, at the top level, we classify errors based on two sources: (1) misprediction of slot-value pairs from the utterances in the current turn $\Delta y^t$ (\textit{delta}), or (2) propagation of mispredictions from past turns $y^{t-1}$ (\textit{error propagation}). These two sources share the same set of specific error types.

At the second level, we classify errors by comparing slot-value pairs between the ground-truth dialogue state and the prediction: (1) a slot-value pair in the ground-truth is absent in the prediction (\textit{miss}), and (2) a slot-value pair in the prediction is absent in the ground-truth (\textit{hallucination}). These two categories contain seven specific error types, detailed as follows.

For the \textit{miss} category, an error may arise because the prediction (1) misses an entire slot-value pair (\textit{miss-total}), (2) misses a special token ``delete'' (\textit{miss-delete}), (3) misses ``don't care,'' which indicates the user's openness to any options (\textit{miss-dontcare}), or (4) misses a slot by confusing it with another slot (\textit{miss-confuse}).

In the \textit{hallucination} category, errors may arise when the prediction (1) hallucinates a slot-value pair that is not informed by the dialogue (\textit{hallucination-total}), (2) hallucinates a value with a correct slot that does not match the dialogue (\textit{hallucination-value}), or (3) incorrectly modifies the previous dialogue state (\textit{hallucination-overwrite}).

This categorization systematically identifies various sources of errors and provides an intuitive framework for understanding DST model weaknesses. Additionally, it highlights the challenges faced by LLMs in key DST capabilities, such as utterance comprehension, hallucination detection, and identifying critical slots. We release the framework as a Python library to facilitate its easy integration in future studies.

\subsection{Error Analysis of Modern LLMs}\label{sec:app_error_analysis}

To understand the weaknesses of modern LLMs in DST, we used our error analysis framework to evaluate the zero-shot and 10-shot DST performance of four models: RefPyDST \citep{king-flanigan-2023-diverse}, \texttt{gpt-3.5-turbo}, \texttt{Llama-3-8B-Instruct}, and the 4-bit quantized \texttt{Llama-3-70B-Instruct} \citep{dubey2024llama}. RefPyDST, the current SOTA model, was reproduced with \texttt{Llama-3-8B-Instruct} \citep{dubey2024llama} because {\texttt{code-davinci-002}} has been deprecated. It uses prompts in Python format, while the other models employ natural language prompts. The prompts used in the experiments are shown in Appendix~\ref{sec:app_prompt}.

In the 10-shot setting, RefPyDST retrieves in-context examples using its own fine-tuned retriever. For the other models, we randomly selected examples from the full training dataset.

Overall, \textit{error propagation} errors were significantly more frequent than \textit{delta} errors. In the zero-shot setting, \textit{error propagation hallucination-total} was the most common, while in the random setting, \textit{error propagation miss-total} dominated. For RefPyDST, an exception was observed where format errors were the most frequent due to Python code formatting, with \textit{hallucination-total} being the next most prominent type of error.

In the 10-shot setting, \textit{miss} errors outnumbered \textit{hallucination} errors in all models except \texttt{gpt-3.5-turbo}. In the zero-shot setting, hallucination errors were generally more prevalent. Across all models, \textit{miss-total} and \textit{hallucination-total} errors—where the correct slot and value are either missed or incorrectly generated—were the most frequent. \textit{Miss-total} errors affected subsequent dialogue turns more extensively, especially for name-related slots, due to issues like user confirmation and flexible preferences not being properly interpreted. Meanwhile, \textit{hallucination-total} errors often stemmed from oversensitivity to system descriptions, incorrect coreference resolution, or user question misinterpretation, notably affecting area-related slots.

To reduce both miss and hallucination errors, addressing \textit{delta} errors and ensuring accurate understanding of current dialogue turns are essential. We now incorporate both the dialogue state and conversation similarity in retrieval and provide explicit instructions in the prompts for better information extraction.

\begin{figure*}[t]
  \includegraphics[width=0.32\linewidth]{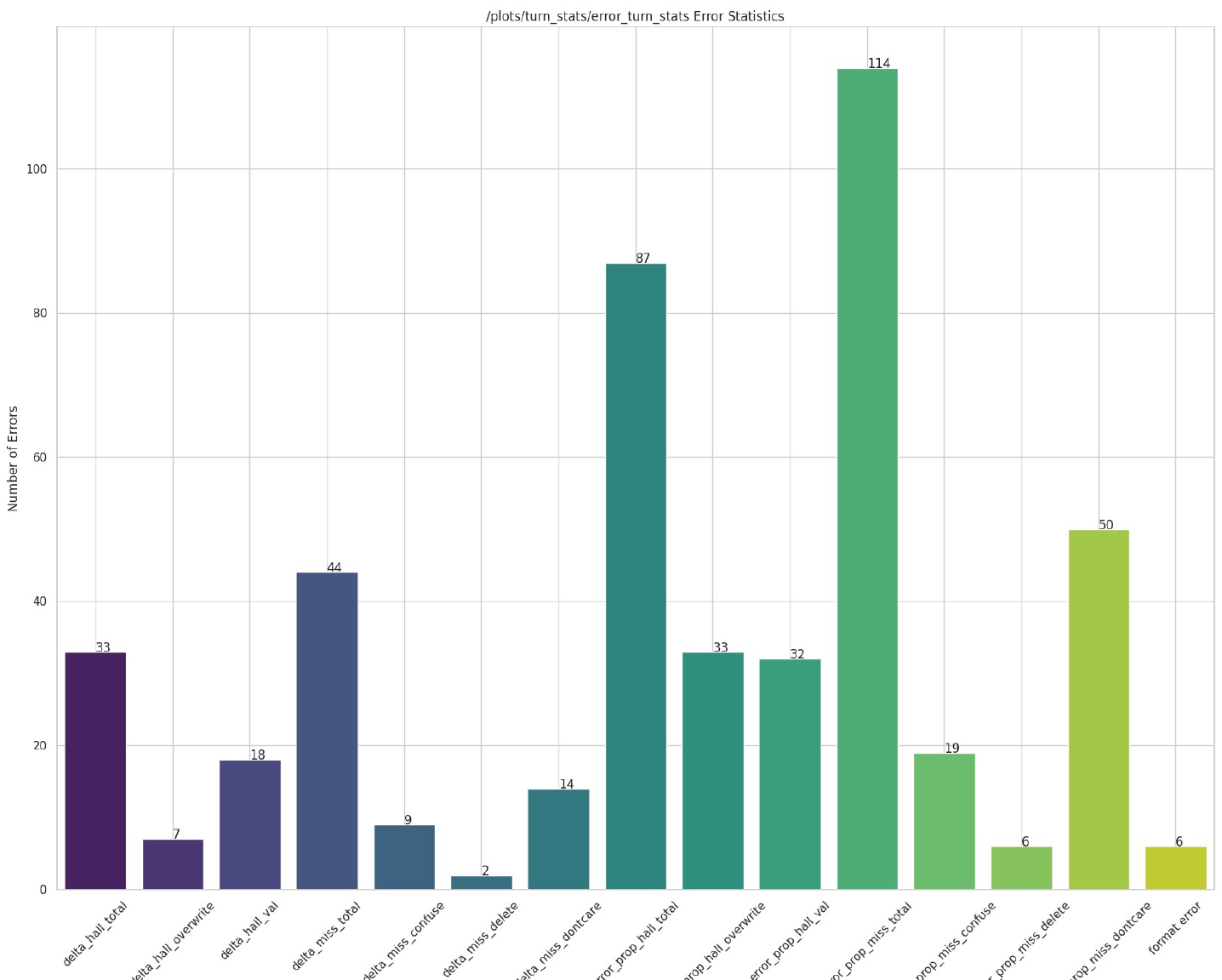} \hfill
  \includegraphics[width=0.32\linewidth]{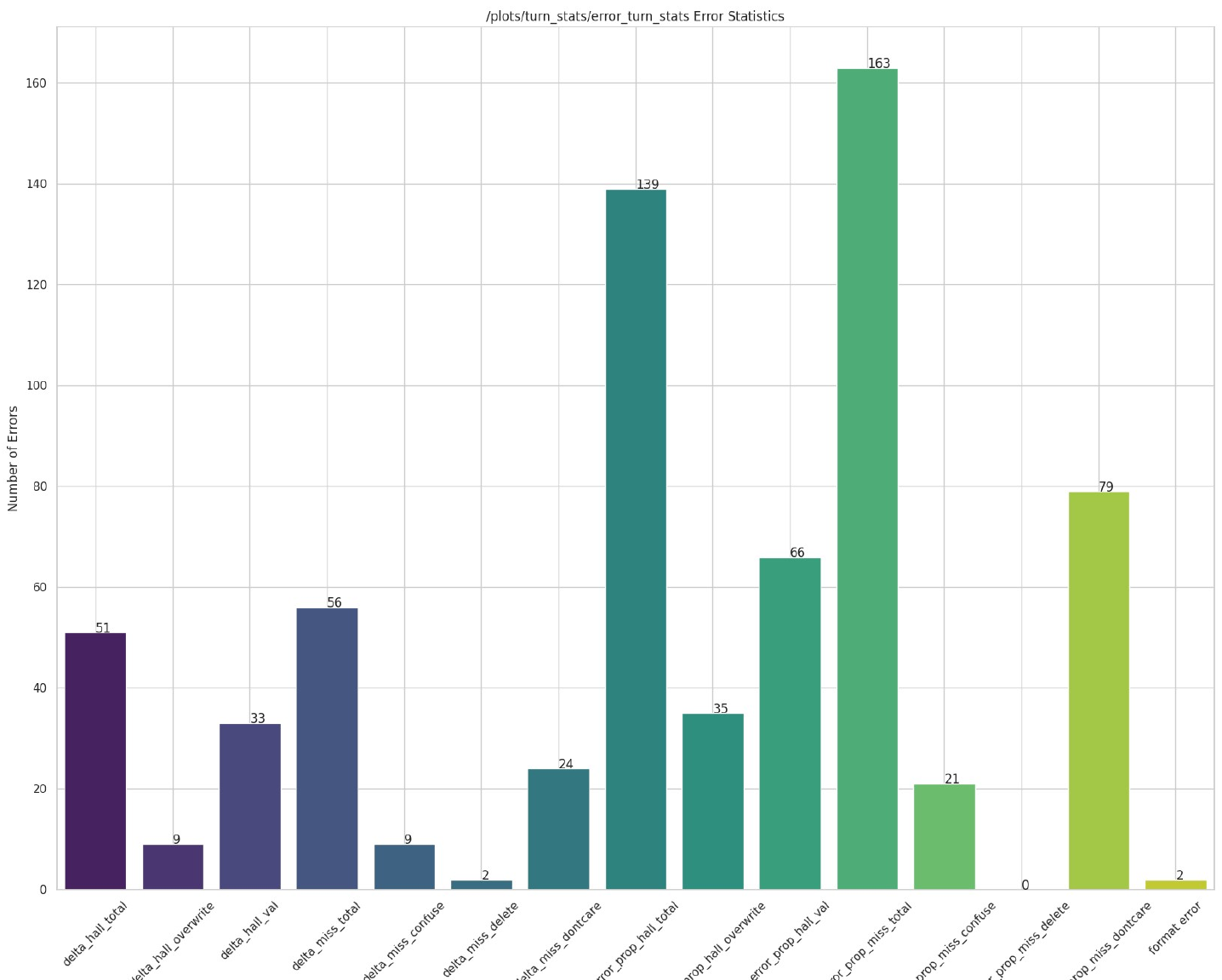} \hfill
  \includegraphics[width=0.32\linewidth]{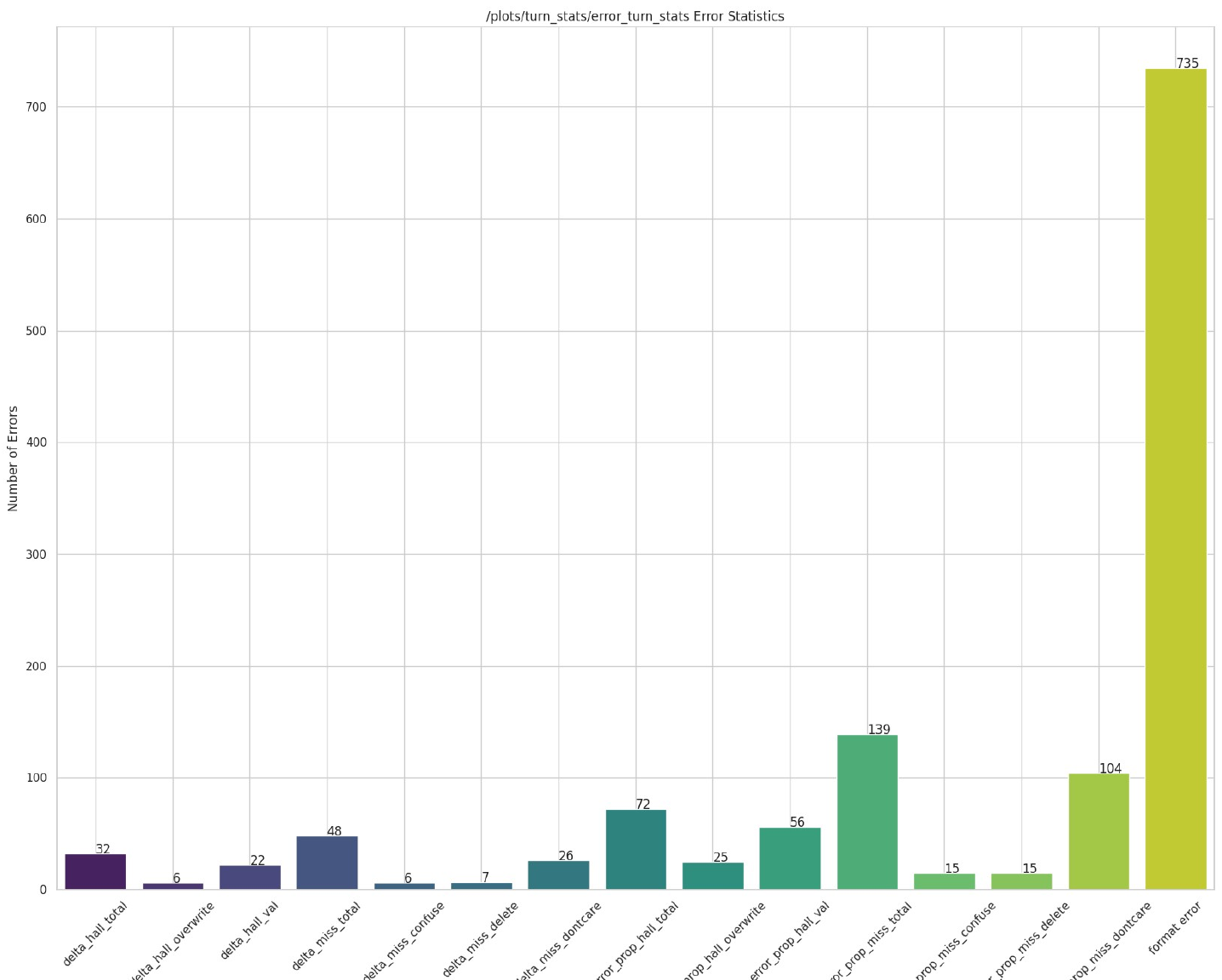}
  \vspace{-0.5em}   \caption {Statistics of error types made by different models in \S{\ref{sec:exp_upperbound}}. From left to right: CombiSearch, Hybrid, and RefPyDST. Formatting errors in RefPyDST are not included.}
  \label{fig:error_types}
\end{figure*}

\section{Experiment Settings}\label{sec:app_exp_settings}

\subsection{Baselines}\label{subsec:app_exp_settings_baseline}
\paragraph{DiSTRICT \citep{venkateswaran-etal-2023-district}} 
Retrieves highly relevant examples based on semantic similarity within dialogues to enhance in-context tuning of T5-small for predicting slot values, eliminating the need for hand-crafted templates.

\paragraph{SGPDST \citep{lee-etal-2021-dialogue}}
Applies schema-driven prompting by providing both dialogue context and schema information into the encoder of a pre-trained T5-base. Its prompts consist of candidate slot labels and their natural language descriptions.

\paragraph{SM2 \citep{chen-etal-2023-stabilized}}
Adapts a meta-ICL framework by training T5-XXL on support dialogue datasets in DST format and evaluating on MultiWOZ 2.4. It aims to stabilize few-shot performance and reduce variance across prompts. It fine-tunes SBERT as an example retriever and uses the predicted dialogue state as a summary of dialogue history.

\paragraph{DS2 \citep{shin-etal-2022-dialogue}} 
Reframes dialogue state tracking as a summarization task, fine-tuning T5-large and BART-large to generate rule-based synthetic summary templates.

\paragraph{IC-DST \citep{hu-etal-2022-context}}
Applies ICL for DST by reformulating the task as a text-to-SQL generation, using SQL to represent dialogue states. An example retriever is trained by scoring examples based on the similarity of their dialogue state changes, and fine-tuning SBERT. Additionally, prior dialogue states are used as context instead of the full dialogue history.

\paragraph{SynthDST \citep{kulkarni-etal-2024-synthdst}} 
Builds on \citet{hu-etal-2022-context}, which reformulated DST as a text-to-SQL task. Unlike \citet{hu-etal-2022-context}, who fine-tuned retrievers for semantically similar dialogues, SynthDST generates high-quality synthetic datasets using LLMs without retriever fine-tuning. With \texttt{code-davinci-002} deprecated, \texttt{gpt-3.5-turbo} is used to maintain system performance.

\paragraph{RefPyDST \citep{king-flanigan-2023-diverse}} 
Follows the ICL framework for DST introduced by \citet{hu-etal-2022-context}. It reformulates DST as a text-to-Python task and trains a retriever using the two-step data generation process from \citet{hu-etal-2022-context}. To enhance inference diversity, it employs maximum marginal relevance for example decoding and PMI-based decoding to mitigate surface form competition. With \texttt{code-davinci-002} deprecated, we replaced it with \texttt{gpt-3.5-turbo} and substituted $\text{PMI}^\beta$ decoding with beam search.

\paragraph{$\text{Se}^2$ \citep{liu-etal-2024-se2}} 
Selects in-context examples sequentially through beam search rather than the traditional `select and organize' approach. This method leverages LLM feedback to capture inter-example relationships and sequential patterns, improving the contextual relevance of ICL prompts. For DST, we applied this approach using Joint F1 as the evaluation metric and \texttt{Llama-3-8B-Instruct} as the LLM to provide task-specific feedback. To reduce computational overhead, we set the beam size to 1 for beam search.

\subsection{Hyperparameters}\label{subsec:app_exp_settings_hyperparam}
All hyperparameter tuning was conducted on the 5\% validation set and performed manually. The parameters used in the CombiSearch process are presented in the ``Common Settings'' paragraph of \S{\ref{sec:exp_settings}}. During inference, we configured both BM25 and SBERT to retrieve the top 10 highest-scoring examples. For LLM inference, beam search was applied as the decoding strategy, with the beam width set to 4. For retriever training, the batch size was set to 32. For each instance, we set the number of hard negatives \( B \) to 16, the number of negatives to 15, and the number of positives to 1. Consequently, each training instance consists of 1 positive, 16 hard negatives, and 15 negatives.

\section{Computing Details}\label{sec:app_exp_settings_computing}

\begin{table}[hbt!]
  \begin{tabular}{>{\centering\arraybackslash}p{2cm}>{\centering\arraybackslash}p{2cm}>{\centering\arraybackslash}p{2cm}}
    \hline
      \textbf{Train Data Size}  &\textbf{\# of instance}     &\textbf{Running Time (h)}\\
      \hline
       1\%      & 524           & 5 \\
       5\%      & 2731          & 26 \\
       100\%     & 54971         & 550   \\
      \hline
  \end{tabular}
  \vspace{-0.5em}   \caption{The number of instances corresponding to 1\%, 5\%, and 100\% training data and the time required to generate \( D_{\text{CombiSearch}} \) by applying CombiSearch to these datasets.}
  \label{tab:combisearch_time}
  \vspace{-1em}  
\end{table}

\begin{table}[hbt!]
  \begin{tabular}{>{\centering\arraybackslash}p{2cm}>{\centering\arraybackslash}p{1.3cm}>{\centering\arraybackslash}p{1.3cm}>{\centering\arraybackslash}p{1.3cm}}
    \hline
      \textbf{Training Method}  &\textbf{1\% (hour)} &\textbf{5\% (hour)} &\textbf{100\% (hour)}\\
      \hline
        CombiSearch             & 4                  & 20                 & 260 \\\hline
        RefPyDST    & 4.5 & 18&36.5               \\
      \hline
  \end{tabular}
  \vspace{-0.5em}   \caption{Total training time for training retrievers based on each method. We report the total time required to train a retriever, so the running time of RefPyDST \citep{king-flanigan-2023-diverse} for 1\% and 5\% is the aggregated training time of the three retrievers.}
  \label{tab:combisearch_retriever_time}
  \vspace{-1em}  
\end{table}

\paragraph{Hardware}  
{All experiments were run on a single NVIDIA A6000 for \texttt{Llama-3-8B-Instruct} and a single NVIDIA A100 for 4-bit–quantized \texttt{Llama-3-70B-Instruct}.  Inference speed was accelerated with vLLM; the 70 B model was additionally compressed with GPTQ (4-bit, group size 128) to fit on one GPU without noticeable accuracy loss.}

\paragraph{Inference cost}  
{Table~\ref{tab:combisearch_time} lists the total wall-clock time required to construct \(D_{\text{CombiSearch}}\) under the 1\%, 5\%, and 100\% training regimes.}

\paragraph{Retriever training}  
{Training is performed on a single NVIDIA A100.  For each dialogue $x_i$ we create 10 positive instances and 31 negatives (§\ref{sec:combisearch_retriever_training}).  Because the large negative pool increases per-epoch cost, we train for 15 epochs on the 1\% and 5\% splits and for 7 epochs on the 100\% split.  Detailed statistics are given in Table~\ref{tab:combisearch_retriever_time}.}

\paragraph{Comparison with RefPyDST.}  
{For the same amount of training data, CombiSearch takes 4.5–16.2x as much time as RefPyDST. The computation time for CombiSearch \textbf{increases linearly with the training data size} as expected, because CombiSearch is not designed to explore all possible combinations. On the other hand, RefPyDST turns out to benefit from a speed boost as training progresses. RefPyDST trains \emph{three} separate retrievers—one per data shard—whereas CombiSearch trains a \emph{single} model on the full split.  This design choice makes CombiSearch slower to train (e.g.\ 260 h vs.\ 36.5 h at 100\%), yet the method still reaches higher Joint Goal Accuracy with only 5\% of the data.  In practice, its stronger generalization shortens the overall \emph{time-to-target-accuracy} despite the longer training phase.
Crucially, however, the extra compute buys markedly better sample efficiency: with only 1\% of the data, CombiSearch already exceeds RefPyDST trained on 5\%, and with open-source inference models it even beats RefPyDST trained on the full corpus. Since pre-training is a one-off expense and retrieval latency at inference time is identical across methods, the longer training phase is quickly amortised; in realistic deployments, months-to-years of serving time dominate the few hundred hours needed to build the CombiSearch retriever, making it the faster route to any target Joint Goal Accuracy.}

\section{Retriever Details}\label{sec:app_ret_details}
In this section, we present the details related to our retriever.  

First, in \S{\ref{subsec:app_ret_details_ret_input_setting}}, we provide a detailed explanation of how each retriever's input is formatted during both training and inference.  
In Appendix~\ref{subsec:app_ret_details_ret_input_comparison}, we provide experimental justification for the approach in which BM25 and SBERT are assigned different input components for the same query, as described in \S{\ref{sec:combisearch}}.

\subsection{Embedding Model Input}\label{subsec:app_ret_details_ret_input_setting}
There are three key stages where the embedding model’s input plays a crucial role: (1) when applying CombiSearch, (2) when training on data obtained from CombiSearch, and (3) when using the trained retriever for inference. We detail how the retriever’s input is structured in each case.  

During the CombiSearch process, all examples are embedded using only their state change information. Specifically, to represent an example \(\{x_i, \Delta y_i\}\), we use only \(\Delta y_i\) for embedding. This applies equally to both training instances and query instances, which are sampled from the training set.

We trained an embedding model on $D_{\text{CombiSearch}}$ to select examples that enhance DST performance when combined in the prompt (Figure~\ref{fig:combisearch}-3) \footnote{We use all-mpnet-base-v2 \cite{song2020mpnetmaskedpermutedpretraining}}.
Each query in $D_{\text{CombiSearch}}$ is paired with a candidate pool of size $N$, where each example is a dialogue context and dialogue state change pair, $\bigl(e_i = (x_i, \Delta y_i)\bigr)$. 
Following prior work \citep{hu-etal-2022-context, king-flanigan-2023-diverse}, we concatenate the dialogue context $y^{t-1}$ and recent utterances $A^t, U^t$ before encoding a turn context $x_i$ (or query) with the embedding model $\textit{emb}$. 
To measure relevance between two input instances $x_i$ and $x_j$, we compute $\cos\bigl(\textit{emb}(x_i), \textit{emb}(x_j)\bigr)$. 

At inference time, we similarly embed a test instance $x_t$ and perform a nearest-neighbor search to retrieve the most relevant examples for the test dialogue context.

\subsection{Optimizing Retriever Inputs for Hybrid Search in CombiSearch}\label{subsec:app_ret_details_ret_input_comparison}
We conducted experiments to determine the optimal input format for each retriever. 

\paragraph{Setting} Few-shot DST was evaluated on a 20\% validation set using 5\% of the training data, assuming oracle settings with access to the gold state. We assessed suitable inputs for BM25 and SBERT in hybrid search (Table~\ref{tab:bm25_input}, Table~\ref{tab:sbert_input}). Based on these results, we selected the two best input formats for each retriever and applied CombiSearch with combinatorial example scoring to evaluate overall performance (Table~\ref{tab:combisearch_ret_input}). Both retriever scores were z-score normalized and multiplied for re-ranking, following the procedure in \S{\ref{sec:exp_settings}}.

\paragraph{Result} As shown in Table~\ref{tab:bm25_input}, BM25 performed best when retrieving examples based only on the slot values of the query's gold state change (\romannumeral 3), followed by using only the utterance and state change slots without history (\romannumeral 2). Similarly, Table~\ref{tab:sbert_input} shows that SBERT achieved the highest performance when retrieving examples solely based on the query's gold state change (\romannumeral 3), with the next best approach incorporating history, utterance, and state change (\romannumeral 1).

To assess hybrid search, we combined the top two input formats from each retriever, resulting in four experimental conditions (Table~\ref{tab:combisearch_ret_input}). The best-performing combination used BM25 with utterance and state change slots (excluding history) (\romannumeral 2) and SBERT with examples retrieved based on the query's gold state change (\romannumeral 3).
We adopted this combination for pool construction in CombiSearch.

\begin{table}[t]
  \begin{tabular}{>{\centering\arraybackslash}p{0.8cm} >{\centering\arraybackslash}p{1cm} >{\centering\arraybackslash}p{1.8cm} >{\centering\arraybackslash}p{0.8cm} >{\centering\arraybackslash}p{0.9cm}}
    \hline
      \textbf{Index}    & \textbf{History}  & \textbf{Recent Utterances}    &\textbf{Gold State}    &\textbf{JGA (\%)}\\
      \hline
       \romannumeral 1  & O                & O                             & $\Delta y_{slot}$             & 55.6            \\
       \romannumeral 2  & X                & O                             & $\Delta y_{slot}$             & 55.7            \\
       \romannumeral 3  & X                & X                             & $\Delta y_{slot}$             & \textbf{66.8}    \\
       \romannumeral 4  & O                & X                             & $\Delta y_{slot}$             & 50.9            \\
      \hline
  \end{tabular}
  \vspace{-0.5em}   \caption{The JGA using BM25 varied its input. `O' indicates inclusion, while `X' indicates exclusion. $\Delta y_{slot}$ represents the slots of state change.}
  \label{tab:bm25_input}
  \vspace{-1em}  
\end{table}

\begin{table}[t]
  \centering
    \begin{tabular}{>{\centering\arraybackslash}p{0.8cm} >{\centering\arraybackslash}p{1cm} >{\centering\arraybackslash}p{1.8cm} >{\centering\arraybackslash}p{0.8cm} >{\centering\arraybackslash}p{0.9cm}}
    \hline
      \textbf{Index} &\textbf{History}  & \textbf{Recent Utterances}    &\textbf{Gold State}    &\textbf{JGA (\%)}\\
      \hline
       \romannumeral 1  & O                & O                             & $\Delta y$             & 44.4            \\
       \romannumeral 2  & X                & O                             & $\Delta y$             & 44.2            \\
       \romannumeral 3  & X                & X                             & $\Delta y$             &\textbf{ 67.3}            \\
       \romannumeral 4  & O                & X                             & $\Delta y$             & 42.0            \\
      \hline
  \end{tabular}
  \vspace{-0.5em}   \caption{The JGA using SBERT varied its input. `O' indicates inclusion, while `X' indicates exclusion. $\Delta y$ represents the state change.}
  \label{tab:sbert_input}
  \vspace{-1em}  
\end{table}

\begin{table}[hbt!]
  \centering
    \begin{tabular}{ccc}
    \hline
      \textbf{Input of BM25}  & \textbf{Input of SBERT}    &\textbf{JGA (\%)}\\
      \hline
       \romannumeral2                & \romannumeral1     & 57.7            \\
       \romannumeral2                & \romannumeral3     & \textbf{64.8}            \\
       \romannumeral3                & \romannumeral1     & 58.8            \\
       \romannumeral3                & \romannumeral3     & 58.8            \\
      \hline
  \end{tabular}
  \vspace{-0.5em}   \caption{The oracle JGA using SBERT and BM25 varied its input. We aggregated the two scores using multiplication. Ultimately, to determine which input to use in the CombiSearch data preparation process, we conducted experiments under the same settings. The Roman numeral (e.g., \romannumeral 1) in the inputs of BM25 and SBERT refers to the index of input combinations presented in the previous tables. Refer to Table~\ref{tab:bm25_input} for BM25 input and Table~\ref{tab:sbert_input} for SBERT input.}
  \label{tab:combisearch_ret_input}
  \vspace{-1em}  
\end{table}

\section{Further Analysis}\label{sec:app_analysis}
Here, we conduct a more in-depth analysis. First, we investigate the effect of the number of in-context examples in Appendix~\ref{subsec:app_analysis_num_shots}. Next, we conduct an experiment by increasing the number of evaluations (\( M \)) for each example in the example scoring process, as described in Appendix~\ref{subsec:app_analysis_num_eval}. Finally, we examine the score aggregation method that integrates the scores returned by BM25 and SBERT in Appendix~\ref{subsec:app_analysis_hybrid_score}.

\subsection{The Effects of the Number of In-Context Examples}\label{subsec:app_analysis_num_shots}
\begin{figure*}[hbt!]
  \includegraphics[width=\textwidth]{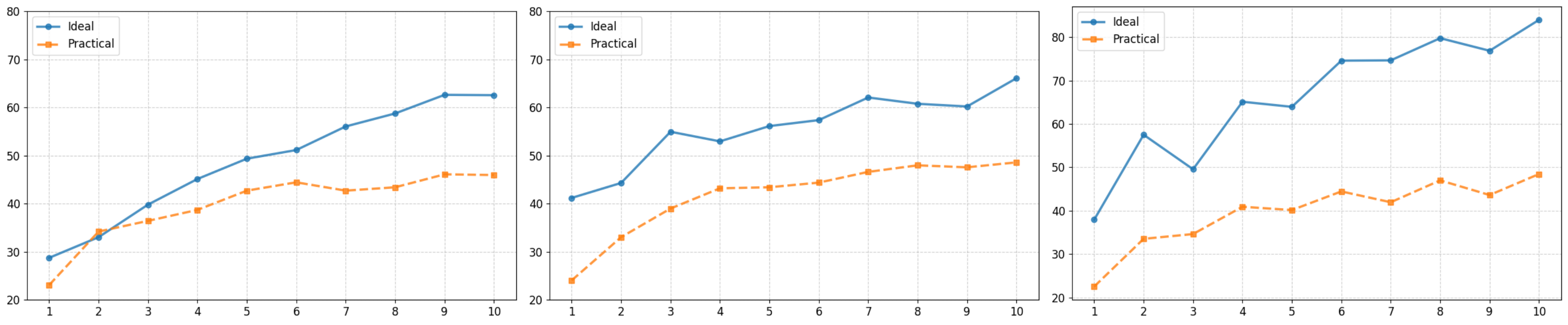}
  \vspace{-0.5em}   \caption{Changes in JGA according to the number of shots. From left to right: BM25, SBERT, and BM25\&SBERT. The ``ideal'' setting refers to a scenario where access to the gold state is available, while the ``practical'' setting refers to a scenario where it is not. Interestingly, when using BM25\&SBERT, the performance gap between the oracle and practical settings was the largest.}
  \label{fig:fig_num_shots}
  \vspace{-1em}
\end{figure*}

Here, we modify the number of in-context examples included in the prompt under both the oracle setting, where the query's gold state change is accessible for retrieval, and the practical setting.

\paragraph{Setting} In both settings, we use 100\% of the training data as the corpus and evaluate performance on a 20\% validation set. We vary the number of in-context examples from 1 to 10 and measure the corresponding performance. We compare BM25, SBERT, and BM25\&SBERT, where SBERT is used without fine-tuning. In the oracle setting, the slot values of the gold state change are used as input for BM25, while SBERT takes the gold state change as input. In the practical setting, both BM25 and SBERT take the dialogue context, represented as the dialogue state and the most recent utterance, as input.

\paragraph{Result} The results in Figure~\ref{fig:fig_num_shots} show that performance tends to improve as more shots are included, without noticeable saturation up to 10 shots. This trend is observed across BM25, SBERT, and BM25\&SBERT. The performance gap between the oracle and practical settings was larger for SBERT than for BM25, suggesting that SBERT is more dependent on access to the gold state. Additionally, SBERT's influence appears to have significantly contributed to the high performance of BM25\&SBERT in the oracle setting.

\subsection{Trade-off Between Accuracy and Cost of the Evaluation Budget $M$}
\label{subsec:app_analysis_num_eval}

{We assess how the per-example evaluation budget \(M\) influences oracle setting performance and compute cost, with the goal of identifying an \(M\) that offers the best balance between JGA and wall-clock time.}

\paragraph{Setting}
{The experiment follows the CombiSearch configuration in
\S\ref{sec:exp_settings} under oracle setting that has access to the gold dialogue state change (oracle mode).  Few-shot DST is evaluated on 1\% and 5\% subsamples of the MultiWOZ training set using a 10\% test slice. The estimated \emph{build time} for the full training set is obtained by multiplying the number of training dialogues by the average latency of scoring each instance $M$ times.}

\paragraph{Results}
{Table~\ref{tab:num_eval} shows that accuracy rises linearly with \(M\), while runtime scales almost linearly.  At 5\% data, increasing \(M\) from~3 to~5 adds
+2.87 absolute point JGA but inflates runtime by 66\% and build cost by 64\%.  With only 1\% data, the same change yields a negligible +0.82\% absolute point. Hence, \textbf{\(M=3\) emerges as the most cost-effective point}, capturing the bulk of the attainable accuracy without incurring prohibitive compute.}

\begin{table}[t]
  \centering
  \small
  \begin{tabular}{>{\centering\arraybackslash}p{1.7cm}%
                  >{\centering\arraybackslash}p{0.7cm}%
                  >{\centering\arraybackslash}p{0.7cm}%
                  >{\centering\arraybackslash}p{1.1cm}%
                  >{\centering\arraybackslash}p{1.6cm}}
    \hline
    \textbf{Evaluations} & \textbf{1\%} & \textbf{5\%} &
    \textbf{Running} & \textbf{Est.\ build}\\
    \textbf{per example} & (JGA) & (JGA) & \textbf{time (h)} & \textbf{time (h)}\\
    \hline
    1  & 65.02 & 69.67 &  2.3 &  180 \\
    2  & 68.98 & 72.81 &  4.2 &  330 \\
    3  & 69.12 & 75.27 &  6.0 &  470 \\
    5  & 69.94 & 78.14 & 10.0 &  769 \\
    10 & 76.91 & 80.87 & 19.5 & 1515 \\
    \hline
  \end{tabular}
  \vspace{-0.5em}
  \caption{{Oracle-mode JGA and compute cost as a function of the evaluation
  budget $M$.  ``Running time'' measures the 10\% test slice; ``Est.\ build
  time'' extrapolates to the full training set.}}
  \label{tab:num_eval}
  \vspace{-1em}
\end{table}


\subsection{Score Aggregation Methods}\label{subsec:app_analysis_hybrid_score}
We compared the performance of two retrievers based on different score aggregation methods. 

\paragraph{Setting} Few-shot DST was conducted using 5\% of the training data and evaluated on three test sets, each comprising 10\% of the full test data. The final performance was determined by averaging the results across these sets. This experiment was conducted after all design choices were finalized to assess the effectiveness of our aggregation method. Since the focus was on evaluating score aggregation, we assumed access to the gold state, eliminating retrieval errors from consideration as in \S{\ref{sec:exp_upperbound}}.

We re-ranked retrieved examples using SBERT and BM25 scores through the following methods:

\paragraph{Multiplication}
The approach selected prior to this experiment, Multiplication, combines SBERT and BM25 scores by multiplying them. Due to scale differences, direct multiplication is infeasible. Instead, both scores are first normalized using z-score transformation before multiplication.

\paragraph{Weighted Sum}
This method sums SBERT and BM25 scores after applying z-score transformation to account for scale differences. Equal weighting is used, setting $\alpha=0.5$.

\paragraph{RRF \citep{10.1145/1571941.1572114}}
Reciprocal Rank Fusion (RRF) assigns a score based on the reciprocal of an example’s rank in each retriever’s list. These scores are then summed across both retrievers. Unlike z-score normalization, RRF does not alter score distributions, making it a widely used and simple re-ranking approach.

\paragraph{Union Top-K}
This method selects the top-K examples by alternating between SBERT and BM25’s ranked lists in a round-robin fashion. If a duplicate appears earlier in the ranking, the next available example is selected instead.

\begin{table}[t]
  \centering
  \begin{tabular}{l|c}
    \hline
      \textbf{Aggregation Method}  & \textbf{JGA (\%)} \\
      \hline
      \textbf{Multiplication}        & \textbf{75.063}    \\
      Weighted Sum          & 71.0      \\
      RRF                   & 70.8      \\
      Union Top K           & 67.7      \\
      \hline
  \end{tabular}
  \vspace{-0.5em}   \caption{The result of score aggregation method.}
  \label{tab:hybrid_score}
  \vspace{-1em}  
\end{table}

\paragraph{Result} Multiplication delivers markedly better performance than the alternative methods, confirming that our design choice was appropriate.

\section{An Example of Retrieved Example}\label{sec:app_coref-ex}

\begin{table*}[t]
    \resizebox{\textwidth}{!}{%
    \centering
    \begin{tabular}{l|r|p{12.0cm}}
        \toprule
        \textbf{Example Type} & \textbf{Category} & \textbf{Example} \\
        \midrule

        \multirow{4}{*}{\textbf{Query}}
            & \textbf{Previous Dialogue State   }& 
                \texttt{restaurant-book day}: saturday, \texttt{restaurant-book people}: 7, \newline
                \texttt{restaurant-book time}: 17:15,\texttt{restaurant-food}: spanish,\newline 
                \texttt{restaurant-name}: la raza, \texttt{restaurant-area}: centre,\newline 
                \texttt{attraction-name}: saint barnabas press gallery \newline\\ 
            & \textbf{System Utterance}          & 
                `saint barnabas press gallery s postcode is cb13ew . \newline would you like more information ?' \newline\\
            & \textbf{User Utterance}            & 
                `no that s it . i also need a taxi taking me from the gallery to the restaurant .' \newline\\
            & \textbf{Dialogue State Change}     & 
                \texttt{taxi-destination}: la raza, \texttt{taxi-departure}: saint barnabas press gallery   \\
        \midrule
        \multirow{4}{*}{\textbf{CombiSearch Example}}
            & \textbf{Previous Dialogue State   }& 
                \texttt{restaurant-book day}: wednesday, \texttt{restaurant-book people}: 6, \newline 
                \texttt{restaurant-book time}: 16:15, \texttt{restaurant-food}: italian, \newline 
                \texttt{restaurant-name}: prezzo, \texttt{restaurant-area}: west, \newline
                \texttt{hotel-book day}: wednesday, \texttt{hotel-book people}: 6,\newline 
                \texttt{hotel-book stay}: 1, \texttt{hotel-name}: huntingdon marriott hotel,\newline 
                \texttt{hotel-area}: west, \texttt{hotel-pricerange}: expensive \newline \\
            
            & \textbf{System Utterance}          & 
                ``i was able to successfully book you for 1 night .\newline  your reference number is bwxli8k3 .   \newline is there anything more i can help you with ?' \newline\\
            & \textbf{User Utterance}            & 
                `i also need a taxi to get from the restaurant to the hotel .' \newline\\
            & \textbf{Dialogue State Change}     & 
                \texttt{taxi-destination}: huntingdon marriott hotel,  \texttt{taxi-departure}: prezzo   \\
        \midrule
        \multirow{4}{*}{\textbf{RefPyDST Example}}
            & \textbf{Previous Dialogue State   }& 
                \texttt{restaurant-name}: pipasha restaurant, \texttt{attraction-name}: club salsa, \newline \\
            
            & \textbf{System Utterance}          & 
                `pipasha is in the east . \newline the full address is newmarket road fen ditton postal code cb58pa . \newline can i help you with anything else ?' \newline\\
            & \textbf{User Utterance}            & 
                `i am gonna need a taxi to get from the club to the restaurant .\newline  can you schedule that for me ?' \newline\\
            & \textbf{Dialogue State Change}     & 
                \texttt{taxi-destination}: pipasha restaurant,  \texttt{taxi-departure}: club salsa  \\
        \bottomrule
    \end{tabular}
    }
        \caption{
        Representative example retrieved by each retriever for the target query. The CombiSearch example shares substantial Previous Dialogue State with the query. This indicates that CombiSearch effectively retrieves examples that share not only dialogue state change but also a similar conversational flow and structural context. 
    }
    \label{tab:coref_example}
\end{table*}

We showcase an example in Table~\ref{tab:coref_example}. We observe substantial overlap between the query’s previous dialogue state and that of the CombiSearch example. Both dialogues contain complex prior states involving two entities---restaurant and attraction in the query, restaurant and hotel in the example---before the user transitions to a new request (booking a taxi). Notably, the CombiSearch example mirrors the query’s dialogue history and therefore presents a user utterance (``i also need a taxi..'') that exhibits a similar structural shift: it refers back to two previously mentioned points of interest (departure and destination) using coreference. Resolving such coreferences is challenging in DST, and in-context examples that display this pattern can enhance model performance. By contrast, although the RefPyDST example also contains a taxi request between two entities, its preceding dialogue state overlaps far less with the query’s rich context and lacks the nuanced history captured by CombiSearch.

This case illustrates that our hybrid CombiSearch retriever finds examples sharing not only dialogue-state similarity but also a parallel conversational flow and structure, leading to significant performance gains.

\section{Prompt for Few-shot DST experiments}\label{sec:app_prompt}

\begin{table*}[t]
\centering
\resizebox{\textwidth}{!}{%
    \begin{tabular}{@{}p{2cm}|p{16cm}@{}}
        \hline
        \textbf{\hspace{2mm}Role} & \textbf{\hspace{8cm}Message} \\ \hline
        \textbf{\hspace{2mm}System} & **Task:** You are an expert in Dialogue State Tracking (DST) focused on managing and updating the dialogue state change based on system-user interactions. The dialogue state represents the user's preferences and booking details across different domains: Hotel, Train, Attraction, Restaurant, and Taxi. \\ \hline
        
        \textbf{\hspace{2mm}User} & 
        **Guidelines:** \newline
        \hspace*{5mm} 1. **Hotel:** \newline
        \hspace*{10mm}- **Slots:** name (string), pricerange (PriceRange), type (HotelType), parking (Option), book stay (integer), book day (DayOfWeek), book people (integer), area (Area), stars (integer between 0 and 5 or "dontcare"), internet (Option). \newline
        \hspace*{10mm}- **Valid Values:** \newline
            \hspace*{15mm}- PriceRange: "dontcare", "cheap", "moderate", "expensive". \newline
            \hspace*{15mm}- HotelType: "hotel", "guest house", "dontcare". \newline
            \hspace*{15mm}- Option: "yes", "no", "dontcare". \newline
            \hspace*{15mm}- DayOfWeek: "monday", "tuesday", "wednesday", "thursday", "friday", "saturday", "sunday". \newline
            \hspace*{15mm}- Area: "dontcare", "centre", "east", "north", "south", "west". \newline\newline
        \hspace*{5mm}2. **Train:** \newline
        \hspace*{10mm}- **Slots:** destination (string), departure (string), day (DayOfWeek), book people (integer), leaveat (hh:mm or "dontcare"), arriveby (hh:mm or "dontcare"). \newline\newline
        \hspace*{5mm}3. **Attraction:** \newline
        \hspace*{10mm}- **Slots:** name (string), area (Area), type (AttractionType). \newline
        \hspace*{10mm}- **Valid Values:** \newline
            \hspace*{15mm}- AttractionType: "architecture", "boat", "church", "cinema", "college", "concert hall", "entertainment", "hotspot", "multiple sports", "museum", "nightclub", "park", "special", "swimming pool", "theatre", "dontcare". \newline\newline
        \hspace*{5mm}4. **Restaurant:** \newline
        \hspace*{5mm}\hspace*{5mm}- **Slots:** name (string), food (string), pricerange (PriceRange), area (Area), book time (hh:mm or "dontcare"), book day (DayOfWeek), book people (integer). \newline\newline
        \hspace*{5mm}5. **Taxi:** \newline
        \hspace*{5mm}\hspace*{5mm}- **Slots:** destination (string), departure (string), leaveat (hh:mm or "dontcare"), arriveby (hh:mm or "dontcare"). \newline\newline
        
        **Example 1 of Dialogue State Change Update Task:**\newline
        \hspace*{5mm}**Previous Belief State:** \newline
        \hspace*{5mm}\hspace*{5mm}\text{\{`train-destination': `cambridge', `train-day': `thursday', `train-arriveby'. `15:45', `train-departure': `kings lynn'\}} \newline\newline
        \hspace*{5mm}**Latest Conversation Between System and User:** \newline
        \hspace*{5mm}\hspace*{5mm}**System:** "It leaves at 14:11 and lasts 47 minutes. Is there anything else?" \newline
        \hspace*{5mm}\hspace*{5mm}**User:** "Yes, I am looking for a restaurant that serves North American food, preferably in the expensive price range."\newline
        
        \hspace*{5mm}**Instructions:**\newline
        \hspace*{5mm}\hspace*{5mm}- Based on the user's latest input, update the belief state by correctly identifying and filling in the relevant domain(s), slot(s) and value(s).\newline
        \hspace*{5mm}\hspace*{5mm}- Provide your output strictly in the Required Output Format below.\newline\
        
        \hspace*{5mm}**Required Output Format:**\newline
        \hspace*{5mm}\hspace*{5mm}**Dialogue state change after Latest Conversation Between System and User:** \newline
        {\hspace*{5mm}\hspace*{5mm}\hspace*{5mm}\text{\{ "DOMAIN\_1- SLOT\_1": "VALUE\_1", "DOMAIN\_2-SLOT\_2": "VALUE\_2", ... \}}} \\ \hline
        
        \textbf{\hspace{2mm}Assistant} &
        {\hspace*{5mm}\hspace*{5mm}\hspace*{5mm}\text{\{`restaurant-food': `north american', `restaurant-pricerange': `expensive'\}}}\\ \hline
    \end{tabular}
}
\vspace{-0.5em}   \caption{An example of prompt for few-shot DST. For brevity, this example only includes one example without query. However, we set ten shots as a default setting for fair comparison with baseline models.}
\label{table:prompt}
\end{table*}

\end{document}